\documentclass[letterpaper, 10 pt, journal, twoside]{IEEEtran}  
\usepackage{amsmath,amsfonts}
\usepackage{algorithmic}
\usepackage{algorithm}
\usepackage{array}
\usepackage[caption=false,font=normalsize,labelfont=sf,textfont=sf]{subfig}
\usepackage{textcomp}
\usepackage{stfloats}
\usepackage{url}
\usepackage{verbatim}
\usepackage{graphicx}
\usepackage{cite}
\usepackage{tabularx}
\usepackage{threeparttable}
\usepackage{multirow} 
\usepackage{color}
\usepackage{gensymb}
\usepackage[section]{placeins}
\usepackage{float}
\usepackage{makecell}
\usepackage[switch]{lineno}
\usepackage{booktabs}
\usepackage[table]{xcolor}
\usepackage{pifont}

\begin{document}

\title{Physics-informed Deep Mixture-of-Koopmans Vehicle Dynamics Model with Dual-branch Encoder for Distributed Electric-drive Trucks}

\author{
Jinyu Miao$^{1}$, Pu Zhang$^{2}$, Rujun Yan$^{1}$, Yifei He$^{1}$, Bowei Zhang$^{1}$, Zheng Fu$^{1}$, Ke Wang$^{2}$, Qi Song$^{1}$, \\ Kun Jiang$^{1,*}$, Mengmeng Yang$^{1}$, Diange Yang$^{1,*}$ 
\thanks{This work was supported in part by the National Natural Science Foundation of China (52394264, 52472449, 52372414, 52402499), and China Postdoctoral Science Foundation (2024M761636).}
\thanks{$^{1}$Jinyu Miao, Rujun Yan, Yifei He, Bowei Zhang, Zheng Fu, Qi Song, Kun Jiang, Mengmeng Yang, and Diange Yang are with the School of Vehicle and Mobility, and State Key Laboratory of Intelligent Green Vehicle and Mobility, Tsinghua University, Beijing, China.{\tt\small jinyu.miao97@gmail.com}}%
\thanks{$^{2}$Pu Zhang and Ke Wang are KargoBot Inc., Beijing, China.}%
\thanks{$^{*}$Corresponding author: Diange Yang, Kun Jiang}
}



\maketitle

\begin{abstract}

Advanced autonomous driving systems require accurate vehicle dynamics modeling. However, identifying a precise dynamics model remains challenging due to strong nonlinearities and the coupled longitudinal and lateral dynamic characteristics.
Previous research has employed physics-based analytical models or neural networks to construct vehicle dynamics representations. Nevertheless, these approaches often struggle to simultaneously achieve satisfactory performance in terms of system identification efficiency, modeling accuracy, and compatibility with linear control strategies.
In this paper, we propose a fully data-driven dynamics modeling method tailored for complex distributed electric-drive trucks (DETs), leveraging Koopman operator theory to represent highly nonlinear dynamics in a lifted linear embedding space. To achieve high-precision modeling, we first propose a novel dual-branch encoder which encodes dynamic states and provides a powerful basis for the proposed Koopman-based methods entitled KODE. A physics-informed supervision mechanism, grounded in the geometric consistency of temporal vehicle motion, is incorporated into the training process to facilitate effective learning of both the encoder and the Koopman operator. Furthermore, to accommodate the diverse driving patterns of DETs, we extend the vanilla Koopman operator to a mixture-of-Koopman operator framework, enhancing modeling capability.
Simulations conducted in a high-fidelity TruckSim environment and real-world experiments demonstrate that the proposed approach achieves state-of-the-art performance in long-term dynamics state estimation.

\end{abstract}


\section{Introduction}
\label{sec:intro}

Safety-guarantee autonomous driving requires precise identification of vehicle dynamics \cite{Jin2019,zhang2024}. Many well-established controllers, such as model predictive control (MPC) \cite{Moser2018} and linear quadratic regulators (LQR) \cite{Chen2019}, rely on accurate dynamics models to achieve high-performance trajectory tracking. However, constructing an accurate vehicle dynamics model remains technically challenging \cite{spielberg2019neural}, particularly for distributed electric-drive trucks (DETs), which feature higher centers of gravity (CoG) and more complex control flexibility compared to traditional internal combustion engine-drive vehicles (ICVs) \cite{guo2024}.

\begin{figure}[!t]
    \centering
    \includegraphics[width=0.97\linewidth]{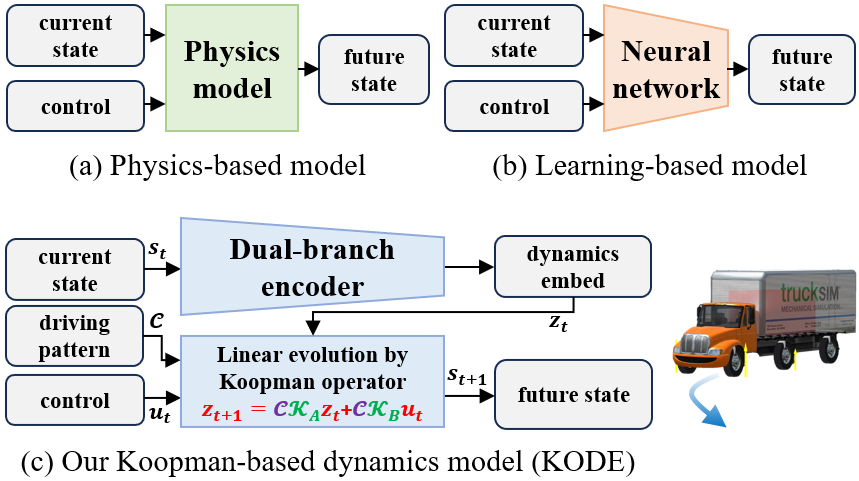}
    \caption{A diagram of (a) physics-based dynamics model, (b) learning-based dynamics model, and (c) the proposed Koopman-based dynamics model KODE using a dual-branch encoder and MoK operator, respectively.}
    \label{fig:intro}
\end{figure}

Over the past decades, researchers have pursued accurate vehicle dynamics modeling. Classical physics-based methods derive analytical formulations to mathematically describe vehicle behavior \cite{Kapania,li2015,Tekin2010,Azizul2024,Pacejka01011992}. However, their accuracy is inherently limited by prior knowledge of the real vehicle, and their generalizability is often compromised by linearization and simplifying assumptions \cite{Subosits2021}. Moreover, their performance depends heavily on identified system parameters, which is costly and labor-intensive \cite{Vicente2021,Johan2019}.
In response to these limitations, recent research has shifted toward learning-based approaches, where deep neural networks are employed as data-driven alternatives to analytical models \cite{spielberg2019neural,Chrosniak2024}, as compared in Fig. \ref{fig:intro} (a) and (b). These methods identify system (network) parameters directly from data, thereby significantly reducing the need for parameter calibration \cite{Xu2019}. Nevertheless, the nonlinearity and lack of interpretability in neural network-based models hinder integration with linear controllers \cite{Moser2018,Chen2019}. This limits their practical deployment in safety-critical autonomous driving systems, where controller transparency and reliability are paramount.

Koopman operator theory \cite{Mauroy2020} provides a powerful framework for analyzing nonlinear systems by lifting states into an infinite-dimensional embedding space, and has recently been applied to vehicle dynamics modeling. To obtain a acceptable finite-dimensional yet accurate linear representation, dimensionality reduction techniques such as dynamic mode decomposition (DMD) \cite{SCHMID2010} and extended DMD (EDMD) \cite{EDMD} have been proposed to approximate the Koopman operator. These approaches typically rely on manually designed kernel functions for state lifting, which is non-trivial for highly nonlinear systems and consequently limits modeling performance. To address this, DeepEDMD \cite{DeepEDMD} and deep direct Koopman (DDK) \cite{DDK} introduce multi-layer perceptron (MLP) encoders and enable pure data-driven system identification via gradient backpropagation. Thanks to the linear structure in the lifted embedding space, Koopman-based models can be readily integrated with linear controllers \cite{Moser2018,Chen2019}. Despite these advances, existing Koopman-based methods have been validated only on small or regular ICVs. Their applicability to DETs, which exhibit significantly higher system complexity, remains unexplored and poses new challenges for accurate dynamics state estimation.

To address these challenges, we propose an enhanced \textbf{KO}opman-based dynamics modeling framework tailored for high-precision state estimation towards \textbf{DE}Ts, entitled KODE. As shown in Fig. \ref{fig:intro}, our improvements span three aspects: encoder architecture, training supervision, and Koopman operator formulation. Specifically, a novel dual-branch encoder combining Transformer and MLP modules is designed to embed dynamics states. A mixture-of-Koopman (MoK) operators is introduced, where the appropriate operator is selected based on the current driving pattern. The encoded dynamics embeddings are evolved linearly via the selected Koopman operator and subsequently decoded to predict future states. Additionally, a geometric consistency loss is incorporated during training to promote long-term prediction accuracy. Benefiting from these innovations, the proposed method achieves precise modeling of the complex dynamics of DETs.
The primary contributions in this work are summarized as follows:
\begin{itemize}
\item To the best of our knowledge, this work is the first to introduce Koopman-based dynamics modeling to distributed electric trucks characterized by expanded control space and complex dynamical behavior, and to achieve high-fidelity dynamics analysis for such platforms.
\item We propose three key enhancements to Koopman-based vehicle dynamics modeling:
    \begin{itemize}
    \item[$\Diamond$] A novel dual-branch encoder that lifts states into high-dimensional embeddings, providing a powerful basis for our Koopman-based dynamics model;
    \item[$\Diamond$] A physics-informed loss function that enforces geometric consistency in the vehicle temporal motion, thereby improving long-term estimation fidelity and physical correlation between various states;
    \item[$\Diamond$] A mixture-of-Koopman (MoK) operator framework that assigns specialized Koopman operators to distinct driving patterns, enhancing modeling accuracy across diverse patterns.
    \end{itemize}
These components efficiently enhance the capability of Koopman-based vehicle dynamics models without requiring extra training data or incurring significant parameter overhead, and they are universally applicable as plug-and-play improvements for any Koopman-based method.
\item We collect vehicle dynamics data for DETs from both high-fidelity simulations and real-world tests. Comprehensive experiments demonstrate that the proposed method achieves state-of-the-art (SOTA) performance in dynamics state estimation.
\end{itemize}

\section{Related Works}
\label{sec:review}

\subsection{Physics-based Vehicle Dynamics Model}

Traditional physics-based methods model vehicle dynamics through mathematical formulations, spanning a range of complexities and precisions. The simplest models neglect vehicle dimensions and load transfer, treating the vehicle as an infinitesimal point mass \cite{Kapania, Subosits2019}. Three degree-of-freedom (DoF) bicycle models introduce geometric realism and capture the three fundamental DoF of vehicle motion, enabling widespread application in planning and control \cite{Timings2013, Liu2018}. To capture additional dynamical effects, such as yaw moments arising from differential drive and brake forces across an axle, which simpler models fail to represent \cite{Subosits2021}, more complex dynamics models have been developed, including 7 DoF \cite{li2015}, 8 DoF \cite{Tekin2010}, and 14 DoF \cite{Azizul2024} models, as well as the well-known Pacejka tire model \cite{Pacejka01011992}.
However, these approaches suffer from two inherent limitations. First, physics-based modeling necessarily relies on local linearization and simplifying assumptions, yet real vehicles are complex electro-mechanical systems rather than ideal rigid bodies. This discrepancy inherently constrains both the accuracy and generalization capability of physics-based models. Second, these models depend on physical parameters that need to be identified, which is called system identification process \cite{Vicente2021,Johan2019}. While existing methods employ Gaussian processes \cite{Kabzan2019} or neural networks \cite{COSTA2023104469} for parameter identification, certain parameters like cornering stiffness coefficients remain inherently unmeasurable \cite{Vicente2021}. These challenges are particularly pronounced for DETs, whose dynamical systems are substantially more complex than those of conventional ICVs.

\subsection{Learning-based Vehicle Dynamics Model}

Inspired by recent advances in deep learning, researchers have introduced neural networks to learn vehicle dynamics directly from data, circumventing the need for traditional analytical formulations. Spielberg \textit{et al.} experimentally demonstrated that deep neural networks can serve as effective dynamics models, often outperforming physics-based approaches \cite{spielberg2019neural}. The Baidu Apollo team further showed that such data-driven solutions substantially reduce development effort by eliminating the need for explicit analytical modeling and system identification, while maintaining high accuracy \cite{Xu2019}. As a result, learning-based vehicle dynamics models using deep neural networks have been widely adopted and further developed in subsequent years \cite{Hermansdorfer2020, Cao2021}.
Some methods integrate physics-based models with neural networks, where the networks are designed to estimate model parameters rather than directly regressing dynamics states \cite{Kim2022, Chrosniak2024}. This hybrid approach effectively mitigates the drift and poor generalization commonly associated with purely data-driven models. Other researchers have adopted a complementary strategy, employing neural networks to compensate for the estimation errors of physics-based methods \cite{DRC-baidu, DRC-baidu2}. This emerging paradigm achieves excellent accuracy in dynamics state estimation, even for complex platforms such as distributed electric vehicles \cite{miao2025}.
Despite their success in high-precision dynamics modeling, neural network-based approaches suffer from limited interpretability and inherent nonlinearity, which hinder their integration into well-posed model-based linear controllers, such as MPC \cite{Moser2018} and LQR \cite{Chen2019}.

\subsection{Koopman-based Vehicle Dynamics Model}

The inherent nonlinearity of real vehicle dynamics poses a fundamental challenge for linear controllers, which require linear system representations. To address this issue, the Koopman operator \cite{Mauroy2020} has been introduced into vehicle dynamics modeling. Researchers have developed several approaches to approximate finite-dimensional Koopman operators, including DMD \cite{SCHMID2010} and its variants such as EDMD \cite{Williams2015,Mamakoukas2021,EDMD} and kernel-based DMD \cite{Matthew2015}, which lift dynamics states into high-dimensional embedding spaces using kernel functions. However, the performance of these methods critically depends on the selection of appropriate kernels, which requires domain expertise and may not optimally capture the intricate nonlinearities inherent in complex vehicle dynamics.
To overcome this limitation, neural networks have been adopted as learnable kernel functions, enabling automatic learning of both network parameters and the Koopman operator in a pure data-driven manner. This paradigm reduces manual effort in kernel design and system identification while achieving improved modeling performance \cite{Lusch2018,DeepEDMD,SafEDMD}. For instance, DeepEDMD \cite{DeepEDMD} introduces multi-step supervision during training to enhance long-term prediction accuracy. Building on this foundation, Xiao \textit{et al.} learn Koopman eigenvalues to construct a block-diagonal transition matrix as the Koopman operator \cite{DDK}, while Shi \textit{et al.} incorporate an auxiliary control network to encode nonlinear state-dependent control components, effectively modeling input nonlinearities \cite{DK-U}.

However, existing Koopman-based methods have been primarily validated on small or regular ICVs where simple state encoders, regression-based supervision, and single Koopman operator suffice to achieve satisfactory performance. When applied to DETs with significantly increased system complexity, the performance of vanilla Koopman operators remains suboptimal, leaving substantial room for improvement. This gap motivates the work presented in this paper.

\section{Methodology}
\label{sec:method}

In this section, we first formalize the dynamics state estimation problem. Next, we review the fundamentals of Koopman operator theory. Subsequently, we present the core contributions of this work, followed by a description of the simulation data collection pipeline used for model training and evaluation. Finally, we introduce a sim-to-real vehicle adaptation strategy to deploy the proposed method on a real-world truck.

\subsection{Problem Statement}
\label{sec:problem}

This paper considers a class of DETs characterized by a six-wheel, three-axis configuration and all-wheel steering functionality.
Denote $\textbf{s}={\left[p_x~p_y~\alpha_z~v_x~v_y~w_z\right]}\in \mathbb{R}^{N_s}$ and $\textbf{u}={\left[T^f_{l}~T^f_{r}~T^m_{l}~T^m_{r}~T^r_{l}~T^r_{r}~\delta^f_{l}~\delta^f_{r}~\delta^m_{l}~\delta^m_{r}~\delta^r_{l}~\delta^r_{r}\right]}\in \mathbb{R}^{N_c}$ as the considered dynamics states and control inputs of the DETs, respectively, where $p_x,p_y$ are the CoG position of trucks in global frame, $\alpha_z$ is the heading/yaw angle, $v_x,v_y,w_z$ are the longitudinal and lateral CoG velocities and heading angular velocity in ego frame, $T^{f/m/r}_{l/r}$ is the driving torque applied to the left (l) or right (r) wheel on the front (f), middle (m) or rear (r) axis, and $\delta$ is the steering angle. Thus, a vehicle dynamics model can be mathematically described as a mapping function $\mathcal{F}_{\theta}:\mathbb{R}^{N_s} \times \mathbb{R}^{N_c} \rightarrow \mathbb{R}^{N_s}$:
\begin{equation}
    \label{equ:1}
    \dot{\textbf{s}}=\mathcal{F}_{\theta}(\textbf{s},\textbf{u})
\end{equation}
where $N_s=6,N_c=12$ in this paper, $\theta$ is the system parameters of the dynamics model. 
In practical applications, we often adopt a discrete-time version of Eqn \ref{equ:1}:
\begin{equation}
    \label{equ:2}
    \textbf{s}_{t+1|t}=\mathcal{F}_{\theta}(\textbf{s}_{t},\textbf{u}_t)
\end{equation}
where $t$ in the index of discrete timestamp. And the one-step estimation results from $t$-th timestamp to $t+1$ timestamp can be simplify as $\textbf{s}_{t+1}=\textbf{s}_{t+1|t}$. For the ease of reading, we format the long-horizon estimation process as:
\begin{equation}
\begin{aligned}
    \label{equ:3}
    \textbf{s}_{t+H_e|t}&=\mathcal{F}_{\theta}(\cdots\mathcal{F}_{\theta}(\textbf{s}_{t},\textbf{u}_t),\cdots,\textbf{u}_{t+H_e-1}) \\
    &=\mathcal{F}_{\theta}(\textbf{s}_{t},\textbf{u}_{t:t+H_e-1})
\end{aligned}
\end{equation}
where $H_e$ is the estimation horizon. 

The task of vehicle dynamics state estimation aims to constuct a reasonable model $\mathcal{F}_{\theta}$ that can accurately govern the state transition of the vehicle over time as a function of control inputs:
\begin{equation}
    \label{equ:4}
    \min_{\mathcal{F},\theta} ||\mathcal{F}_{\theta}(\textbf{s}_{t},\textbf{u}_t)-\textbf{s}_{t+1}||
\end{equation}
However, accurate modeling of real-world vehicle dynamics remains technically challenging due to the strong coupling between longitudinal and lateral dynamics, as well as the nonlinear characteristics of tires and suspensions. On one hand, the inherent linearity and simplifying assumptions required in conventional modeling approaches inevitably compromise model fidelity. On the other hand, identifying the system parameters $\theta$ is both costly and difficult. These limitations underscore the need for a novel methodology that can simultaneously achieve high-fidelity vehicle dynamics modeling and enable efficient system identification.

\begin{figure*}[!t]
    \centering
    \includegraphics[width=0.97\linewidth]{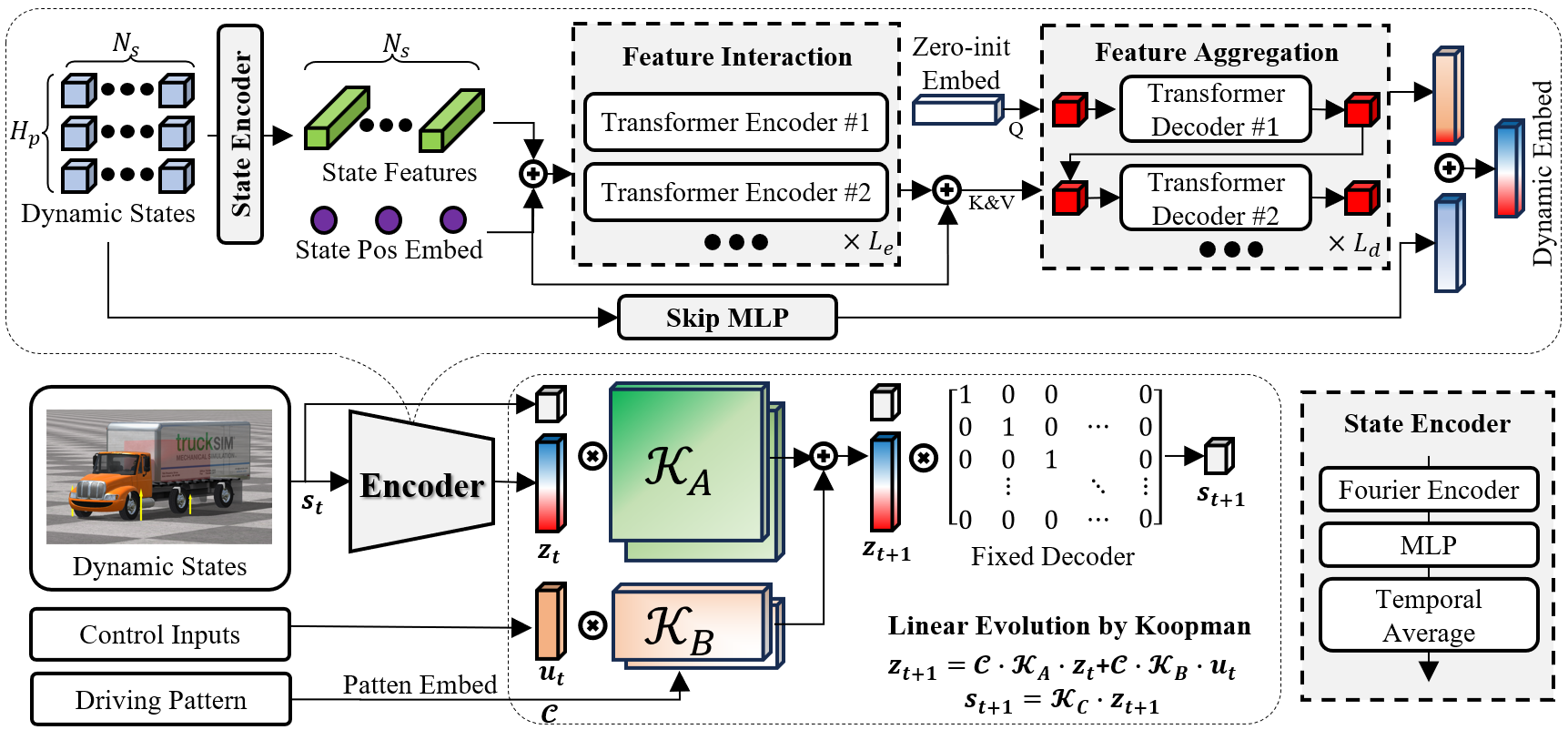}
    \caption{An overview of the proposed method. The dynamics states are encoded to dynamics embeddings through a dual-branch encoder. Specific Koopman operator is selected by given driving pattern. Then, the dynamics embeddings are linearly evolved and decoded to future dynamics states.}
    \label{fig:overview}
\end{figure*}

\subsection{Preliminaries: Koopman Operator for Forced Systems}

The Koopman operator provides a powerful operator-theoretic framework for modeling nonlinear dynamical systems. Rather than propagating states directly in the original low-dimensional space, this operator acts on a space of observable functions, enabling linear dynamical evolution in an infinite-dimensional Hilbert space for unforced nonlinear systems \cite{Lusch2018,Takeishi2017,Brunton2022}. By incorporating control inputs into the observable functions, the Koopman framework can be extended to forced systems given in Eqn. \ref{equ:1}:
\begin{equation}
    \label{equ:5}
    g(\textbf{s}_{t+1},\textbf{u}_{k+1})=\mathcal{K}\cdot g(\textbf{s}_{t},\textbf{u}_{k})
\end{equation}
where $g:\mathbb{R}^{N_s+N_c}\rightarrow\mathbb{R}^\infty$ is the encoding function that lifts the state space to a embedding space, and $\mathcal{K}$ represents the Koopman operator, which is typically an infinite-dimensional linear operator in Hilbert space. For practical implementation, it is necessary to design appropriate finite-dimensional encoding functions to obtain a tractable approximation of the Koopman operator. Generally, A common simplification is to partition the encoding function such that control inputs are not encoded: $g(\textbf{s}_{t},\textbf{u}_{k})=\left[\textbf{z}_t;\textbf{u}_{k}\right]$, where $\textbf{z}_t$ represents the encoded state embeddings. Then, Eqn. \ref{equ:5} can be simplified as:
\begin{equation}
    \label{equ:6}
    \left[\begin{array}{c}
         \textbf{z}_{t+1}  \\
         \textbf{u}_{t+1} 
    \end{array}\right]=\left[\begin{array}{cc}
         \mathcal{K}_A & \mathcal{K}_B  \\
         \mathcal{K}_* & \mathcal{K}_*
    \end{array}\right] \left[\begin{array}{c}
                                 \textbf{z}_{t}  \\
                                 \textbf{u}_{t} 
                            \end{array}\right]
\end{equation}
Since the future control signals $\textbf{u}_{t+1}$ are determined by the controller, the dynamics model based on $D$-order Koopman operator can be expressed as:
\begin{equation}
    \label{equ:7}
    \textbf{z}_{t+1}=\mathcal{K}_A\cdot \textbf{z}_{t}+\mathcal{K}_B\cdot\textbf{u}_{t}
\end{equation}
where $\mathcal{K}_A \in \mathbb{R}^{D\times D},\mathcal{K}_B\in \mathbb{R}^{D \times N_c}$ are operator parameters.

Following prior work \cite{DDK,DeepEDMD,DK-U,ESO-DK}, the encoded emebddings $\textbf{z}_{t}$ typically  consist of the original dynamics states and features extracted by an encoder $\mathcal{E}(\cdot)$:
$\textbf{z}_{t}=\left[\textbf{s}_{t};\mathcal{E}(\textbf{s}_{t})\right]$.
After linear evolution in embedding space, the future dynamics states can be recovered by a decoder: $\textbf{s}_{t+1}=\mathcal{D}(\textbf{z}_{t+1})$.

The impracticality of infinite-dimensional embeddings in real-world settings demands effective encoder-decoder designs for finite-order Koopman-based dynamics models. Early implementations utilized manually designed kernels for encoding and decoding \cite{EDMD}, with least-squares estimation for $\mathcal{K}_A,\mathcal{K}_B$. Later, deep learning-enabled Koopman models replaced these with trainable MLP encoders and decoders, optimized via gradient backpropagation \cite{EDMD,DeepEDMD}. Although successful for traditional ICVs, these approaches struggle with DETs due to their expanded state and control dimensionality. We attribute this limitation to three key factors: insufficient encoder capability, inadequate training supervision, and ill-suited Koopman operator modeling.

\subsection{KODE: Deep Koopman with Dual Branch Encoder}
\label{sec:network}

In this work, we first introduce a novel dual-branch dynamics encoder that maps dynamics states into expressive embeddings, providing a powerful foundation for high-precision Koopman-based dynamics modeling, as in Fig. \ref{fig:overview}. 
The first branch leverages a Transformer-based encoder and decoder for rich representation learning, while the second branch comprises an MLP module to ensure stable training convergence.

We consider a $H_p$-step temporal dynamics sequence $\textbf{s}_{t-H_p:t}\in \mathbb{R}^{N_s\times (H_p+1)}$, with $t$ representing the index of current timestamp. The significant scale variations across different dynamics states, especially vehicle positions, hinder raw state encoding by neural networks. Conventional methods normalize states based on predefined ranges. However, such approaches fail in unbounded open scenarios where range priors are unavailable. We therefore convert the position components of the temporal dynamics states into a local ego frame centered at the current timestamp $t$ via $\textbf{s}_{t-H_p:t}[:2]=\textbf{s}_{t-H_p:t}[:2]-\textbf{s}_{t}[:2]$
such that the ranges of different states become comparable and bounded, facilitating effective learning by neural networks.

{
\setlength{\parindent}{0cm}
\textbf{Feature Extraction:} 
The processed temporal dynamics states are fed into a state encoder for initial encoding. Each state is treated as an independent element and first encoded using Fourier position encoding \cite{tancik2020fourier} to preserve subtle state variations. These encoded features are then passed through a two-layer MLP with GeLU activations \cite{gelu} and averaged along the temporal dimension:
}
\begin{equation}
    \textbf{f}_t=\texttt{mean}(\texttt{MLP}(\texttt{Fourier}(\textbf{s}_{t-H_p:t}))) \in \mathbb{R}^{N\times D}
\end{equation}
Here, various dynamics states are separately encoded and temporally aggregated to high-dimensional state features.

\begin{figure}[!t]
    \centering
    \includegraphics[width=0.97\linewidth]{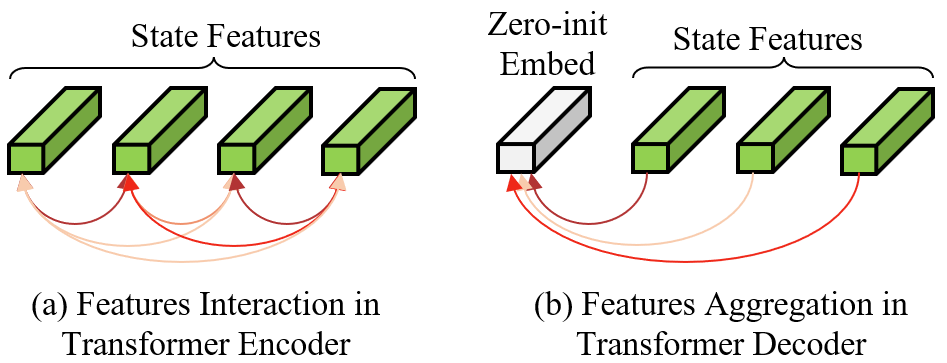}
    \caption{A diagram of feature interaction by Transformer encoder and feature aggregation by Transformer decoder, respectively.}
    \label{fig:transformer}
\end{figure}

{
\setlength{\parindent}{0cm}
\textbf{Feature Interaction:} 
To obtain expressive encoded state features, modeling interactions between different states is essential, as it enables the synthesis of composite features. Therefore, following individual state encoding, we introduce a Transformer encoder module for state-wise interaction. 
}
\begin{equation}
    \textbf{f}^e_t=\texttt{encoder}(\textbf{f}_t,\textbf{pos}) \in \mathbb{R}^{N\times D}
\end{equation}
A learnable state position embedding $\textbf{pos}\in \mathbb{R}^{N_s\times D}$ is added to the state features to distinguish different states. As illustrated in Fig. \ref{fig:transformer}, the self-attention mechanism in the Transformer encoder facilitates global interaction among dynamics states, allowing each feature to aggregate information from others in proportion to their relevance, thereby constructing a representative basis for the embedding space.

{
\setlength{\parindent}{0cm}
\textbf{Feature Aggregation:} 
Unlike common feature aggregation methods that simply average or summarize features, we instead employ a Transformer decoder-based module to aggregate the $N_s$ state features into the final Transformer embeddings:
}
\begin{equation}
    \textbf{z}^{\text{Trans}}_t=\texttt{decoder}(q=\textbf{z}_0,k=v=\textbf{f}^e_t,\textbf{pos}) \in \mathbb{R}^{D}
\end{equation}
where $\textbf{z}_0\in \mathbb{R}^{D}$ is a zero-initialized learnable query, and the state position embedding $\textbf{pos}$ is shared between the feature interaction and aggregation stages, serving as a unique identifier for each dynamics state. 
As illustrated in Fig. \ref{fig:transformer}, this module fuses multiple dynamics features into a single embedding by computing adaptive weights based on the relevance of each feature. 
Intuitively, the cross-attention mechanism in the Transformer decoder assigns importance weights to different basis in the embedding space and aggregates them to produce the final representation. 
This approach is conceptually similar to the class token in Transformer-based image classification \cite{vit} and the pose query in Transformer-based pose estimation \cite{poet}, both of which have demonstrated significant success, validating the effectiveness of this aggregation strategy.

{
\setlength{\parindent}{0cm}
\textbf{Skip Connection:} 
To ensure the training stability of deep neural networks, we add a shallow MLP module to directly encode dynamics states and obtain a MLP embedding:
}
\begin{equation}
    \textbf{z}^{\text{MLP}}_t=\texttt{MLP}(\left[\textbf{s}_{t-H},\textbf{s}_{t-H+1},\cdots,\textbf{s}_{t}\right]) \in \mathbb{R}^{D}
\end{equation}
where $\left[\cdots\right]$ is channel-wise concatenation.

Finally, the final dynamics embedding is obtained by:
\begin{equation}
    \textbf{z}_t=\textbf{z}^{\text{Trans}}_t+\textbf{z}^{\text{MLP}}_t \in \mathbb{R}^{D}
\end{equation}
After linear evolution using the Koopman operator, we use a fixed decoder $\mathcal{K}_C$ to recover the dynamics states from the dynamics embedding:
\begin{equation}
\begin{aligned}
    \textbf{s}_{t+1}&=\mathcal{K}_C\cdot\textbf{z}_{t+1} \\
    \mathcal{K}_C&=\left[\begin{array}{cc}
        \texttt{diag}(N) & 0 \\
        0 & 0
    \end{array}\right] \in \mathbb{R}^{N\times D}
\end{aligned}
\end{equation}
where $\texttt{diag}(N)$ is a $N$-order diagonal matrix.
Note that for fair comparison with existing methods, the historical state length $H$ is typically set to 0 (no historical states are used).

\subsection{Physics-informed Learning for Deep Koopman}
\label{sec:loss}

We adopt a multi-objective training framework for our data-driven Koopman dynamics model, incorporating standard state and feature losses \cite{EDMD,DeepEDMD} alongside a novel physics-informed geometry consistency loss.

{
\setlength{\parindent}{0cm}
\textbf{Multiple-step Estimation:} 
As demonstrated in \cite{DeepEDMD}, incorporating multi-step supervision improves long-term estimation fidelity. Given the linear evolution structure of the Koopman framework, the $H_e$-step state estimation in Eqn. \ref{equ:3} can be written as:
}
\begin{equation}
\begin{aligned}
    \textbf{s}_{t+H_e|t}&=\mathcal{F}_\theta(\textbf{s}_t,\textbf{u}_{t:t+H_e-1}) \\
                      &=\mathcal{K}_C\cdot\textbf{z}_{t+H_e|t} \\
    \textbf{z}_{t+H_e|t}&=\mathcal{K}_A\cdot\textbf{z}_{t+H_e-1|t}+\mathcal{K}_B\cdot\textbf{u}_{t+H_e-1} \\
                      &=\mathcal{K}^2_A\cdot\textbf{z}_{t+H_e-2|t}+\mathcal{K}_A\mathcal{K}_B\cdot\textbf{u}_{t+H_e-2}+\mathcal{K}_B\cdot\textbf{u}_{t+H_e-1} \\
                      &=\cdots \\
                      &=\mathcal{K}^{H_e}_A\cdot\textbf{z}_{t}+\sum^{H_e}_{i=1}\mathcal{K}^{i-1}_A\mathcal{K}_B\cdot\textbf{u}_{t+H_e-i}
\end{aligned}
\end{equation}

In this work, we employ both single-step ($H_e=1$) and multi-step ($H_e=100$) supervision to comprehensively train the dynamics model for optimal performance.

{
\setlength{\parindent}{0cm}
\textbf{State Loss:} First, we measure the estimation errors in original dynamics state space as state loss $\mathcal{L}_{s}$:
}
\begin{equation}
    \mathcal{L}_{s}=\underbrace{\frac{1}{H_e}\sum^{H_e}_{i=1}{||\textbf{s}_{t+i}-\hat{\textbf{s}}_{t+i}||}_2}_{\text{single-step}}+\underbrace{\frac{1}{H_e}\sum^{H_e}_{i=1}{||\textbf{s}_{t+i|t}-\hat{\textbf{s}}_{t+i}||}_2}_{\text{multi-step}}
\end{equation}
where $\hat{\textbf{s}}$ is the ground-truth dynamics states.

{
\setlength{\parindent}{0cm}
\textbf{Feature Loss:} Then, to ensure the capability of linear evolution in lifted embedding space, we measure the estimation errors in dynamics embedding space as feature loss $\mathcal{L}_{f}$:
}
\begin{equation}
    \mathcal{L}_{f}=\frac{1}{H_e}\sum^{H_e}_{i=1}{||\textbf{z}_{t+i}-\hat{\textbf{z}}_{t+i}||}_2+\frac{1}{H_e}\sum^{H_e}_{i=1}{||\textbf{z}_{t+i|t}-\hat{\textbf{z}}_{t+i}||}_2
\end{equation}
where $\hat{\textbf{z}}$ is the dynamics embedding of ground-truth states.

\begin{figure}[!t]
    \centering
    \includegraphics[width=0.97\linewidth]{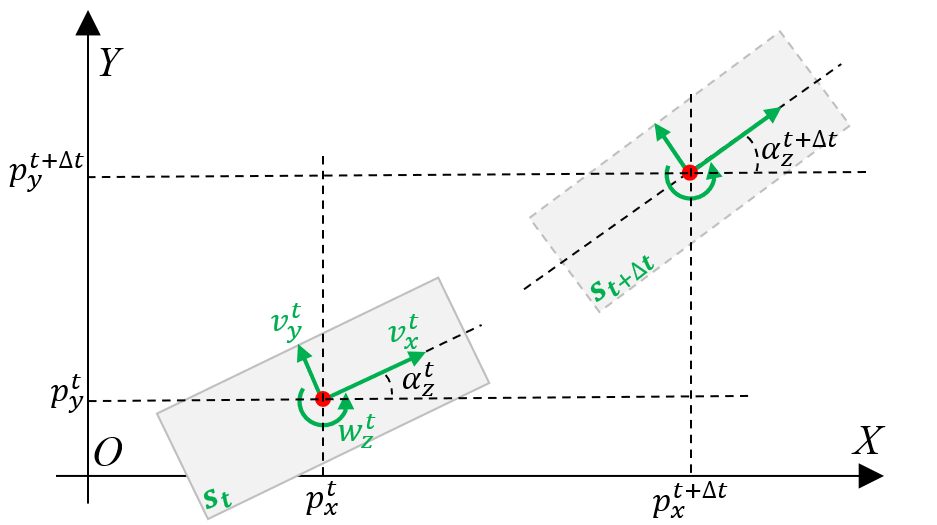}
    \caption{A diagram of geometric consistency of the temporal vehicle motion.}
    \label{fig:geometry}
\end{figure}

{
\setlength{\parindent}{0cm}
\textbf{Physics-informed Geometry Consistency Loss:} 
Although the state loss and feature loss enable the Koopman-based dynamics model to progressively learn state estimation in a time-series regression manner, they overlook the geometric relationships among different dynamics states along the temporal dimension. As illustrated in Fig. \ref{fig:geometry}, the temporal motion of a vehicle inherently follows the geometric principles of rigid-body motion. Consequently, the evolution of vehicle poses, specifically global position $p_x, p_y$ and heading angle $\alpha_z$, can be approximately derived from linear velocities $v_x, v_y$ and angular velocity $w_z$ as follows:
}
\begin{equation}
\begin{aligned}
    \label{equ:geometry}
    \dot{p}^t_x&=\lim_{\Delta t\rightarrow 0}\frac{p^{t+\Delta t}_x-p^t_x}{\Delta t}=v^t_x\cdot \texttt{cos}(\alpha^t_z)-v^t_y\cdot\texttt{sin}(\alpha^t_z) \\
    \dot{p}^t_y&=\lim_{\Delta t\rightarrow 0}\frac{p^{t+\Delta t}_y-p^t_y}{\Delta t}=v^t_x\cdot \texttt{sin}(\alpha^t_z)+v^t_y\cdot\texttt{vos}(\alpha^t_z) \\
    \dot{\alpha}^t_z&=\lim_{\Delta t\rightarrow 0}\frac{\alpha^{t+\Delta t}_z-\alpha^t_z}{\Delta t}=w^t_z \\
\end{aligned}
\end{equation}

Based on this principle, we propose a geometric consistency loss. First, we measure the discrepancy between estimated poses and those derived from velocity estimates over time. 
\begin{equation}
\begin{aligned}
    \delta^t_x&=\left|\frac{p^{t+\Delta t}_x-p^t_x}{\Delta t}-(v^t_x\cdot \texttt{cos}(\alpha^t_z)-v^t_y\cdot\texttt{sin}(\alpha^t_z))\right| \\
    \delta^t_y&=\left|\frac{p^{t+\Delta t}_y-p^t_y}{\Delta t}-(v^t_x\cdot \texttt{sin}(\alpha^t_z)+v^t_y\cdot\texttt{cos}(\alpha^t_z))\right| \\
    \delta^t_\alpha&=\left|\frac{\alpha^{t+\Delta t}_z-\alpha^t_z}{\Delta t}-w^t_z\right| \\
\end{aligned}
\end{equation}

In practice, the discrepancy cannot be eliminated entirely due to measurement noise and the discrete-time interval $\Delta t$. We therefore compute ``reference'' discrepancy $\hat{\delta}^t_x,\hat{\delta}^t_y,\hat{\delta}^t_\alpha$ from the ground-truth states to establish tolerance thresholds. 
As different driving conditions and patterns yield varying scales of discrepancy, we adaptively normalize the estimated discrepancy against these tolerances, yielding the geometric consistency errors for position and orientation:
\begin{equation}
    e^t_x=\frac{{\left(\delta^t_x-\hat{\delta}^t_x\right)}^+}{\texttt{max}~\hat{\delta}^*_x+\epsilon},
    e^t_y=\frac{{\left(\delta^t_y-\hat{\delta}^t_y\right)}^+}{\texttt{max}~\hat{\delta}^*_y+\epsilon},
    e^t_\alpha=\frac{{\left(\delta^t_\alpha-\hat{\delta}^t_\alpha\right)}^+}{\texttt{max}~\hat{\delta}^*_\alpha+\epsilon}
\end{equation}
where ${\left(\cdot\right)}^+$ truncates the negative values (\textit{i.e.}, when the estimated geometric consistency discrepancy is lower than the ground-truth value), and $\epsilon=1e{}^{-6}$ is a tiny constant to avoid numerical  division by zero.

To prioritize long-term estimation accuracy, we introduce a temporal weighting scheme that assigns increasing weight $r$ to geometric consistency errors at later timestamps:
\begin{equation}
    r_t=1+{\left(\frac{t}{L}\right)}^2
\end{equation}

Therefore, the proposed physics-informed geometric consistency loss $\mathcal{L}_{g}$ can be formatted as:
\begin{equation}
    \mathcal{L}_{g}=\frac{1}{L}\sum^L_{t=1}r_t\cdot\left(e^t_x + e^t_y + \lambda\cdot e^t_\alpha\right)
\end{equation}
where $\lambda$ is a weighting factor.

In all, the overall loss function used in this work is written as:
\begin{equation}
    \label{equ:final}
    \mathcal{L}=\mathcal{L}_s+\beta\mathcal{L}_f+\eta\mathcal{L}_g
\end{equation}
where $\beta,\eta$ are weighting factors.

\begin{figure}[!t]
    \centering
    \includegraphics[width=0.97\linewidth]{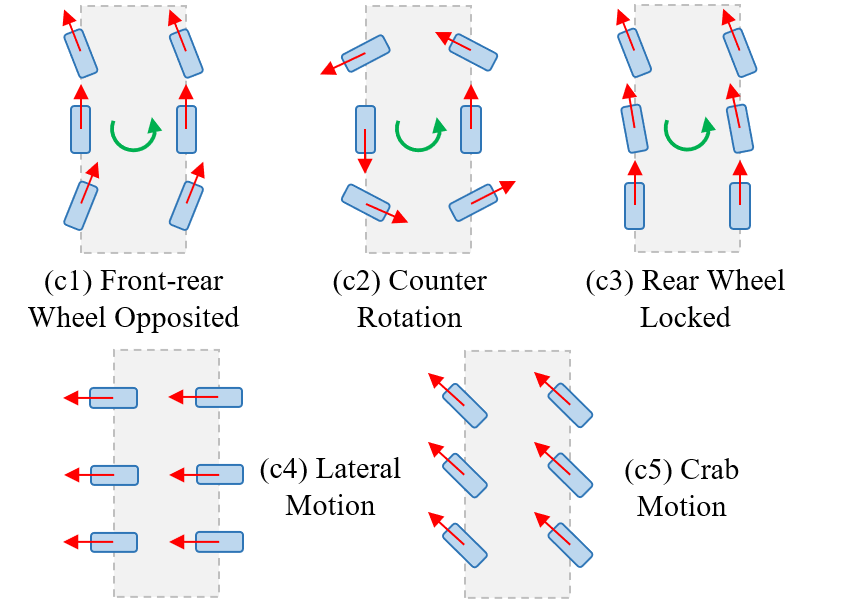}
    \caption{A diagram of five driving patterns in this work.}
    \label{fig:pattern}
\end{figure}

\subsection{Mixture-of-Koopmans for Varying Driving Patterns}
\label{sec:moe}

Each wheel in DETs can be independently driven and steered, resulting in an action space of dimensionality $\mathbb{R}^{M}$, which is substantially larger than that of conventional ICVs. This expanded action space introduces significant complexity for both dynamics modeling and control. In practice, the motion of DETs can be categorized into five typical driving patterns based on the characteristics of the applied control actions: front-rear wheel opposite (FRO), counter rotation (COR), rear-wheel locked (RWL), lateral motion (LAM), and crab motion (CRM). Formal definitions of these patterns are provided in Table \ref{table:patterns}, with visual illustrations in Fig. \ref{fig:pattern}. Among these, FRO, RWL, and CRM are commonly used during normal operation, whereas COR and LAM rarely occur unless required for specialized maneuvers such as U-turns or parallel parking. For safety reasons, pattern switching is not allowed in a single driving maneuver. 
Consequently, the collected data from both simulation and real-world exhibits significant imbalance across patterns. 

In this work, we observe that a single, unified Koopman operator fails to generalize well across all patterns due to both the distinct control characteristics and the imbalanced training data. To this end, we propose a MoK operator inspired by the mixture-of-experts framework \cite{moe}, where each ``expert'' corresponds to a Koopman operator specialized for a specific driving pattern. This approach enhances modeling capacity without requiring additional training data for rare patterns or increasing the dimension of dynamics embeddings.

The implementation is straightforward. Formally, we define a set of Koopman operators $\{\mathcal{K}^i_A\}^5_{i=1} \in \mathbb{R}^{5\times D\times D},\{\mathcal{K}^i_B\}^5_{i=1}\in \mathbb{R}^{5\times D\times M}$ as the MoK operator. The driving pattern associated with each run is encoded as a one-hot embedding $\mathcal{C}\in\mathbb{R}^5$. During dynamics state estimation, the specific Koopman operator corresponding to the current driving pattern is selected. Accordingly, Eqn. \ref{equ:7} can be reformulated as:
\begin{equation}
    \textbf{z}_{t+1}=\mathcal{C}\cdot\mathcal{K}_A\cdot \textbf{z}_{t}+\mathcal{C}\cdot\mathcal{K}_B\cdot\textbf{u}_{t}
\end{equation}

In addition, we also propose a weighting scheme for the MoK operator based on the proportion of each driving pattern:
\begin{equation}
    \kappa(\mathcal{C}) =\texttt{log}(\sum_i V_i+1) /  \texttt{log}({V_c+1})
\end{equation}
where $V_c$ is the amount of training data using driving pattern $c$. The weight is multiplied by the batch-wise training loss:
\begin{equation}
    \mathcal{L}_{batch}=\frac{\sum_{b} \kappa(\mathcal{C}_b)\cdot\mathcal{L}_b}{\sum_{b} \kappa(\mathcal{C}_b)}
\end{equation}
where $b$ is the batch index.
This scheme helps the MoK operator to emphasize samples using rare driving conditions.

\begin{table*}[!t]
	\centering
	\caption{The definition of five driving patterns for distibuted electric drive trucks.}
    \renewcommand\arraystretch{1.5}
	\begin{threeparttable}
    \resizebox{\linewidth}{!}{
		\begin{tabularx}{\linewidth}{ l l l | X | p{4.0cm} X }
			\toprule
			{Pattern} & train  & val & Definition & Data Generation Rules \\
			\midrule
            c1:FRO & 154.3K & 13.9K & The rear wheels are steered in the opposite direction to the front wheels, reducing the turning radius. Under high-speed steering in simulation, this pattern may induce rollover incidents where the wheels on one side lift off the ground. & $T^{*}_{*}=T,~\delta^f_{*}=-\delta^r_{*}=\delta,~\delta^m_{*}=0$ \\
            c2:COR & 9.2K & 2.3K& This pattern, also termed the ``tank turn'' pattern, steers the left and right, front and rear wheels in opposite directions, aligning all steering centers at the vehicle's CoG. It permits rotation around the vertical axis with no translational movement. & $T^f_{*}=-T^m_{l}=T^m_{r}=T^r_{*}=T,~\delta^f_{l}=-\delta^r_{l}=\pi-\delta^f_{r}=\delta^r_{r}+\pi=\delta>\pi/2,~\delta^m_{*}=0$ \\
            c3:RWL & 92.0K & 18.4K & The front wheels and middle wheels are steered according to Ackerman steering geometry, whereas the rear wheels are locked with no steering input. & $T^*_{*}=T, ~\delta^f_{*}=\delta, ~\delta^m_{*}=\texttt{atan}({\delta}/{2}), ~\delta^r_{*}=0$ \\
            c4:LAM & 11.5K & 2.3K & All wheels are steered to 90$\degree$ relative to the vehicle's longitudinal axis, enabling lateral movement for parallel parking or precise positioning in confined spaces. & $T^*_{*}=T,~\delta^*_{*}=\delta=\pi/2$ \\ 
            c5:CRM & 92.0K & 18.4K & All wheels are oriented at an identical angle and direction, producing diagonal movement without altering the vehicle's heading angle. & $T^*_{*}=T,~\delta^*_{*}=\delta<\pi/2$ \\
			\bottomrule
		\end{tabularx}
    }
	\end{threeparttable}
	\label{table:patterns}
\end{table*}

\begin{table}[!t]
	\centering
	\caption{The configuration of the truck in the simulation.}
    \renewcommand\arraystretch{1.5}
	\begin{threeparttable}
    \resizebox{\linewidth}{!}{
		\begin{tabular}{r r | r r }
			\toprule
			{Config} & Value & Config & Value \\
			\midrule
            sprung mass & 4455.0 kg & load mass & 5000.0 kg \\
            unsprung mass & 1050.0 kg & track width & 2.030 m \\
            axis f-m distance & 3.250 m & axis f-r distance & 6.500 m \\
			\bottomrule
		\end{tabular}
    }
	\end{threeparttable}
	\label{table:config}
\end{table}

\subsection{Vehicle State Data Collection by Simulation}
\label{sec:data}

Given the lack of publicly available dynamics datasets for DETs, we conduct simulations to generate the necessary data for validating our method. Following common practice in autonomous driving research, we employ a six-wheel DET model with independent steering capabilities in the TruckSim simulator. Detailed conguration settings are listed in Table \ref{table:config}. Steering angles and driving torques are generated using sinusoidal waveforms to simulate a wide range of driving conditions, including acceleration, deceleration, constant-speed driving, straight-line motion, and turning, following \cite{miao2025}. To ensure the validity of the control inputs, all generated sequences comply with the constraints of the five driving patterns and are manually verified. Given that COR and LAM patterns are rare in practice, we collect fewer samples for these patterns to reflect real-world usage. Additionally, we simulate high-speed steering scenarios using FRO pattern that leads to rollover, to improve the model's generalization capability. The final dataset comprises approximately 414,400 valid samples, split into training and validation sets as detailed in Table \ref{table:patterns}.

\subsection{Sim-to-Real Vehicle Adaptation}
\label{sec:transfer}

According to Koopman operator theory, the operators $\mathcal{K}_A$ and $\mathcal{K}_B$ serve as the system parameters of the dynamical system in the high-dimensional embedding space. Consequently, this formulation enables a practically feasible and efficient adaptation strategy: given a well pre-trained  encoder, identifying a new vehicle reduces to learning only the Koopman operator. In this work, we pre-train the encoder using large-scale, high-quality, and diversely distributed simulation data, enabling it to effectively lift raw dynamics states into expressive embeddings. When deploying the proposed method to real-world applications, we therefore freeze the pre-trained encoder and adapt only the Koopman operator. To mitigate overfitting on limited real-world data, we further introduce a regularization loss $\mathcal{L}_r$:
\begin{equation}
    \mathcal{L}_r={||\mathcal{K}_A||}_2 + {||\mathcal{K}_B||}_2
\end{equation}
and add this item into the training loss in Eqn. \ref{equ:final} with a tiny weighting factor.

\section{Experiments}
\label{sec:result}

\subsection{Experimental Settings}

{
\setlength{\parindent}{0cm}
\textbf{Implementation Details:} 
We implement the proposed Koopman-based dynamics model using the PyTorch library and train it from scratch for 100 epochs on a single NVIDIA 4090D GPU. We employ the ADAM optimizer with a batch size of 512, an initial learning rate of $1\times10^{-3}$, and a weight decay of $1\times10^{-5}$. The learning rate is decayed by a factor of 0.9 every 10 epochs to prevent oscillation and ensure stable convergence. Simulation data are collected using TruckSim 2019.0, where the DET model is developed from a 3A conventional van with six wheels, three axles, and a payload. The weighting factors in the loss function are set as $\lambda=0.01$, $\beta=0.1$, and $\eta=0.05$ across all experiments. By default, the dimension of the lifted dynamics embedding is set to 16 to ensure fair comparison with existing methods \cite{DeepEDMD,DDK}. The depth of Transformer encoders and decoders are set as 2 for efficiency. Control inputs are normalized to the range $[0,1]$. The simulation sampling interval is 10 ms, and we perform state estimation over a horizon of $L=100$ steps, corresponding to 1 second in most experiments.
}

{
\setlength{\parindent}{0cm}
\textbf{Compared Baselines:} 
Due to the lack of publicly available dynamics models for DETs, we reproduce some popular baselines to validate the effectiveness of our method. All approaches under comparison take identical inputs (current dynamics states and control inputs) and estimate the dynamics states for the subsequent timestamp. The evaluated methods span three categories: physics-based analytical models, neural network-based approaches, and Koopman-based methods. Further details are as follows.
}
\begin{itemize}
    \item For physics-based methods, we implement a linear model of the form $\textbf{s}_{t+1}=A\textbf{s}_t+B\textbf{u}_t$ and a 6 DoF model. Their system parameters are either provided by TruckSim (GT), or identified via gradient backpropagation (BP) or least squares solvers (LS).
    \item For neural network-based methods, we adopt commonly used MLP, Transformer \cite{vaswani2017attention}, and the recently proposed KAN \cite{liu2025kan}. In these baselines, states and control inputs are concatenated along the channel dimension and fed into the network. Beyond direct future state estimation, we also implement a variant that estimates the increments ($\Delta$) of dynamics states, which helps mitigate estimation drift and yields improved performance.
    \item For Koopman-based methods, we reproduce EDMD with four different hand-crafted kernel functions \cite{EDMD}, as well as DeepEDMD \cite{DeepEDMD} and DDK \cite{DDK}. Original implementations employ lightweight encoders and rely on pre-defined state ranges for normalization, which leads to suboptimal performance when applied to DETs in open scenarios. To ensure a fair comparison, we adjust the encoder sizes in DeepEDMD and DDK to match the parameter size of our model, and we adopt the same state preprocessing strategy. The modified versions are denoted with an mask (*) for distinction.
\end{itemize}
All compared methods are configured with the same experimental settings as our approach, including the lifted dimension of dynamics embeddings, training epochs, and so on.

{
\setlength{\parindent}{0cm}
\textbf{Evaluation Metrics:} 
The practical utility of a dynamics model lies in its ability to accurately estimate states over extended horizons. Therefore, we assess long-term estimation performance by computing the average trajectory error (MDE) and heading error (MAE) between the estimated and ground-truth states over the estimation horizon:
}
\begin{equation}
\begin{aligned}
    \text{MDE}&=\frac{1}{H_e}\sum^{H_e}_{i=1}{\sqrt{{\left(p^i_x-\hat{p}^i_x\right)}^2+{\left(p^i_y-\hat{p}^i_y\right)}^2}} \\
    \text{MAE}&=\frac{1}{H_e}\sum^{H_e}_{i=1}|\alpha^i_z-\hat{\alpha}^i_z| \\
\end{aligned}
\end{equation}
which reflects the overall accuracy of multi-step dynamics state estimation. Additionally, we measure the final trajectory error (FDE) and heading error (FAE) to analyze the accumulation of estimation errors:
\begin{equation}
\begin{aligned}
    \text{FDE}&=\sqrt{{\left(p^{H_e}_x-\hat{p}^{H_e}_x\right)}^2+{\left(p^{H_e}_y-\hat{p}^{H_e}_y\right)}^2} \\
    \text{FAE}&=|\alpha^{H_e}_z-\hat{\alpha}^{H_e}_z| \\
\end{aligned}
\end{equation}
As heading error is encompassed within trajectory error, we primarily adopt the MDE and FDE metric for evaluation.

\begin{table*}[!t]
	\centering
	\caption{The comparison of 100-step dynamics state estimation with different methods in terms of trajectory errors (meter).}
    \renewcommand\arraystretch{1.5}
	\begin{threeparttable}
    \resizebox{\linewidth}{!}{
		\begin{tabular}{r | r r | r r | r r | r r | r r | r r }
			\toprule
			\multirow{2}{*}{Model} & \multicolumn{2}{c | }{\textbf{c1: FRO}} & \multicolumn{2}{c|}{\textbf{c2: COR}} & \multicolumn{2}{c|}{\textbf{c3: RWL}}  & \multicolumn{2}{c|}{\textbf{c4: LAM}} & \multicolumn{2}{c|}{\textbf{c5: CRM}} & \multicolumn{2}{c}{\textbf{All}} \\
             & MDE $\downarrow$ & FDE $\downarrow$ & MDE $\downarrow$ & FDE $\downarrow$ & MDE $\downarrow$ & FDE $\downarrow$ & MDE $\downarrow$ & FDE $\downarrow$ & MDE $\downarrow$ & FDE $\downarrow$ & MDE $\downarrow$ & FDE $\downarrow$\\
			\midrule
            \multicolumn{12}{l}{\textbf{Physics-based Dynamics methods}} \\
            \hline
			6 DoF model (GT) & 0.0064${}^!$ & 0.0150${}^!$ & \multicolumn{2}{c|}{\textcolor{red}{failed}} & 0.0068 & 0.0162 & \multicolumn{2}{c|}{\textcolor{red}{failed}} & 0.0040 & 0.0097 & 0.0057 & 0.0135 \\
            6 DoF model (BP) & 0.0376 & 0.0977 & 0.0808 & 0.2426 & 0.0425 & 0.1119 & 0.7946 & 2.1898 & 0.0145 & 0.0378 & 0.0648 & 0.1755 \\
            linear model (LS) & 0.9167 & 1.8189 & 0.5379 & 1.0405 & 1.2429 & 2.4641 & 0.1373 & 0.2730 & 0.6020 & 1.2061 & 0.8724 & 1.7331\\
            linear model (BP) & 0.5570 & 1.0847 & 1.8098 & 3.6884 & 1.1240 & 2.1366 & 0.0647 & 0.1564 & 0.1233 & 0.3139 & 0.6329 & 1.2478\\
            \hline
            \multicolumn{12}{l}{\textbf{Pure Neural Network-based Dynamics methods}} \\
            \hline
            MLP & 1.6777 & 3.6579 & 0.0574 & 0.1143 & 1.9547 & 3.9397 & 0.4071 & 0.8423 & 1.5586 & 3.5511 & 1.6101 & 3.4518  \\
            $\triangle$ MLP &  0.0293 & 0.0779 & 0.0535 & 0.1073 & 0.0274 & 0.0712 & 0.0244 & 0.0459 & 0.0128 & 0.0204 & 0.0240 & 0.0564 \\
            KAN \cite{liu2025kan} & 1.6925 & 3.6847 & 0.0764 & 0.1228 & 1.9228 & 3.9090 & 0.3571 & 0.7780 & 1.5945 & 3.6046 & 1.6138 & 3.4638 \\
            $\triangle$ KAN \cite{liu2025kan} &0.0676 &0.1458 & 0.3476 & 0.6888 & 0.0641 & 0.1548 & 0.0809 & 0.1604 & 0.0583 & 0.1354 & 0.0755 & 0.1685 \\
            $\triangle$ Transformer \cite{vaswani2017attention} & 0.0140 & 0.0272 & 0.0421 & 0.0831 & 0.0116 & 0.0235 & 0.0194 & 0.0320 & 0.0110 & 0.0229 & 0.0136 & 0.0271   \\
            \hline
            \multicolumn{12}{l}{\textbf{Koopman-based Dynamics methods}} \\
            \hline
            $\text{EDMD}^*_{\text{Thinplate}}$ \cite{EDMD} & 0.4330 & 0.8552 & 1.2985 & 2.9354 & 0.6774 & 1.3424 & 0.2288 & 0.5310 & 0.2478 & 0.6511 & 0.4802 & 1.0223 \\
            $\text{EDMD}^*_{\text{Gaussian}}$ \cite{EDMD} & 0.5585 & 1.0902 & 1.8374 & 3.7574 & 1.1623 & 2.2602 & 0.0278 & 0.0375 & 0.0895 & 0.1894 & 0.6344 & 1.2468 \\
            $\text{EDMD}^*_{\text{Invquad}}$ \cite{EDMD} & 0.5373 & 1.1711 & 0.5631 & 1.0196 & 0.8221 & 1.6167 & 0.1271 & 0.2625 & 0.2956 & 0.7505 & 0.5357 & 1.1353 \\
            $\text{EDMD}^*_{\text{Invmultquad}}$ \cite{EDMD} & 0.5750 & 1.3703 & 1.1202 & 2.2939 & 0.7544 & 1.5445 & 0.2482 & 0.7975 & 0.3652 & 1.1358 & 0.5740 & 1.3648  \\
            
            DeepEDMD \cite{DeepEDMD} & 0.7465 & 1.1394 & 0.3044 & 0.6822 & 0.7184 & 1.2484 & 0.3297 & 0.6736 & 0.3356 & 0.6122 & 0.5648 & 0.9620 \\
            DeepEDMD* \cite{DeepEDMD} & 0.0334 & 0.0680 & 0.0566 & 0.1012 & 0.0259 & 0.0461 & 0.0051 & \textbf{0.0095} & 0.0346 & 0.0841 & 0.0311 & 0.0650 \\
            DDK* \cite{DDK} & 1.6214 & 3.5264 & 0.0687 & 0.1205 & 1.9247 & 3.8601 & 0.2856 & 0.5447 & 1.2991 & 2.5991 & 1.4950 & 3.0635 \\
            
            \rowcolor{gray!20}KODE (Ours)  & \textbf{0.0120} & \textbf{0.0323} & \textbf{0.0186} & \textbf{0.0331} & \textbf{0.0089} & \textbf{0.0254} & \textbf{0.0051} & 0.0113 & \textbf{0.0055} & \textbf{0.0133} & \textbf{0.0088} & \textbf{0.0228} \\
			\bottomrule
		\end{tabular}
    }
    \begin{tablenotes}
        \footnotesize
        \item[!] The 6 DoF model sometimes fails when the truck is steered in high speed. 
    \end{tablenotes}
	\end{threeparttable}
	\label{table:main-traj}
\end{table*}

\begin{table*}[!t]
	\centering
	\caption{The comparison of 100-step dynamics state estimation with different methods in terms of heading angle errors (degree).}
    \renewcommand\arraystretch{1.5}
	\begin{threeparttable}
    \resizebox{\linewidth}{!}{
		\begin{tabular}{r | r r | r r | r r | r r | r r | r r }
			\toprule
			\multirow{2}{*}{Model} & \multicolumn{2}{c | }{\textbf{c1: FRO}} & \multicolumn{2}{c|}{\textbf{c2: COR}} & \multicolumn{2}{c|}{\textbf{c3: RWL}}  & \multicolumn{2}{c|}{\textbf{c4: LAM}} & \multicolumn{2}{c|}{\textbf{c5: CRM}} & \multicolumn{2}{c}{\textbf{All}} \\
             & MAE $\downarrow$ & FAE $\downarrow$ & MAE $\downarrow$ & FAE $\downarrow$ & MAE $\downarrow$ & FAE $\downarrow$ & MAE $\downarrow$ & FAE $\downarrow$ & MAE $\downarrow$ & FAE $\downarrow$ & MAE $\downarrow$ & FAE $\downarrow$\\
			\midrule
            \multicolumn{12}{l}{\textbf{Physics-based Dynamics methods}} \\
            \hline
			6 DoF model (GT) & 0.0379${}^!$ & 0.0827${}^!$ & \multicolumn{2}{c|}{\textcolor{red}{failed}} & 0.0434 & 0.0956 & \multicolumn{2}{c|}{\textcolor{red}{failed}} & 0.0149 & 0.0331 & 0.0315 & 0.0693 \\
            6 DoF model (BP) & 0.1016 & 0.2786 & 9.5936 & 28.4310 & 0.1134 & 0.3257 & 16.5352 & 45.7721 & 0.0361 & 0.1053 & 1.1613 & 3.2976 \\
            linear model (BP) & 0.3863 & 1.2790 & 2.0805 & 4.0713 & 0.2675 & 0.6080 & 0.1357 & 0.2626 & 0.4337 & 1.4458 & 0.4226 & 1.1852 \\
            \hline
            \multicolumn{12}{l}{\textbf{Pure Neural Network-based Dynamics methods}} \\
            \hline
            $\triangle$ MLP & 2.3063 & 4.6282 & 0.6151 & 1.2107 & 2.3901 & 4.8265 & 0.2186 & 0.4222 & 0.7765 & 1.5661 & 1.6684 & 3.3591 \\
            $\triangle$ KAN \cite{liu2025kan} & 0.6733 & 1.3301 & 8.2556 & 16.4227 & 0.6020 & 1.1366 & 0.6361 & 1.2820 & 0.5764 & 1.1451 & 0.9310 & 1.8295 \\
            $\triangle$ Transformer \cite{vaswani2017attention} & 0.1387 & 0.2185 & 1.5967 & 3.1523 & 0.1604 & 0.2929 & 0.1629 & 0.2829 & 0.1862 & 0.2795 & 0.2233 & 0.3881  \\
            \hline
            \multicolumn{12}{l}{\textbf{Koopman-based Dynamics methods}} \\
            \hline
            $\text{EDMD}^*_{\text{Thinplate}}$ \cite{EDMD} & 1.3388 & 3.2921 & 1.6396 & 3.3221 & 1.7239 & 3.1224 & 0.5773 & 1.0089 & 0.7188 & 1.8324 & 1.2415 & 2.6567 \\
            $\text{EDMD}^*_{\text{Invquad}}$ \cite{EDMD} & 0.4142 & 1.0749 & 0.9720 & 2.0876 & 0.2237 & 0.6498 & 0.4162 & 1.0839 & 0.2774 & 0.4794 & 0.3286 & 0.7780 \\
            
            DeepEDMD* \cite{DeepEDMD} & 1.1743 & 1.6803 & 0.4107 & 0.6030 & 1.0409 & 1.4455 & \textbf{0.0871} & \textbf{0.2832} & 1.0500 & 1.2985 & 1.0117 & 1.3724 \\
            DDK* \cite{DDK} & 20.4075 & 19.9178 & 79.1033 & 81.0054 & 29.0135 & 28.9444 & 3.4534 & 3.4142 & 3.6693 & 3.7408 & 19.4384 & 19.3933 \\
            
            \rowcolor{gray!20}KODE (Ours) & \textbf{0.1435} & \textbf{0.3522} & \textbf{0.3000} & \textbf{0.6591} & \textbf{0.0926} & \textbf{0.2406} & 0.2302 & 0.3334 & \textbf{0.0371} & \textbf{0.0621} & \textbf{0.1013} & \textbf{0.2306} \\
			\bottomrule
		\end{tabular}
    }
	\end{threeparttable}
	\label{table:main-yaw}
\end{table*}

\subsection{Comparison with State-of-the-Art methods}

The proposed method is evaluated against existing baselines on the 100-step dynamics state estimation task. Tables \ref{table:main-traj} and \ref{table:main-yaw} summarize the main results, while further analysis using different driving patterns is also provided.

The 6 DoF model with ground-truth parameters achieves high accuracy on certain driving patterns but fails on others where the operating conditions exceed its design domain. 
When parameters are learned from data, its performance degrades substantially. 
The linear model, despite its widespread use in planning and control, fails to deliver satisfactory modeling accuracy. 
These results highlight the inherent limitation of physics-based approaches that they cannot simultaneously achieve efficient identification, high accuracy, and broad generalization. 
Neural networks \cite{liu2025kan,vaswani2017attention}, while capable of learning nonlinear dynamics from data, suffer from black-box opacity and nonlinearity, which hinder deployment in safety-critical applications and integration with linear controllers \cite{Moser2018,Chen2019}. 
Koopman-based methods offer a compelling alternative by lifting nonlinear dynamics into a linear embedding space compatible with linear control frameworks. They also enable data-driven parameter learning, circumventing costly system identification. 
Among these, our approach, KODE, consistently outperforms all baselines across diverse driving patterns.

Experimental results further reveal that manually designed kernels \cite{EDMD} produce poor lifted embeddings, leading to unsatisfactory accuracy. DeepEDMD \cite{DeepEDMD} replaces fixed kernels with learnable MLP encoders, but its original normalization scheme fails in unbounded scenarios. With our preprocessing, its performance improves markedly, surpassing EDMD by a large margin and confirming the value of learnable encoders. DDK \cite{DDK} exhibits convergence issues in our setting. By contrast, our proposed method, KODE, which integrates a dual-branch neural network-based encoder with the Koopman framework and achieves enhancements from both physics-informed supervision and the MoK operator consistently outperforms all Koopman-based baselines, demonstrating its superior representation capacity. 

\begin{figure}[!t]
    \centering
    \includegraphics[width=0.97\linewidth]{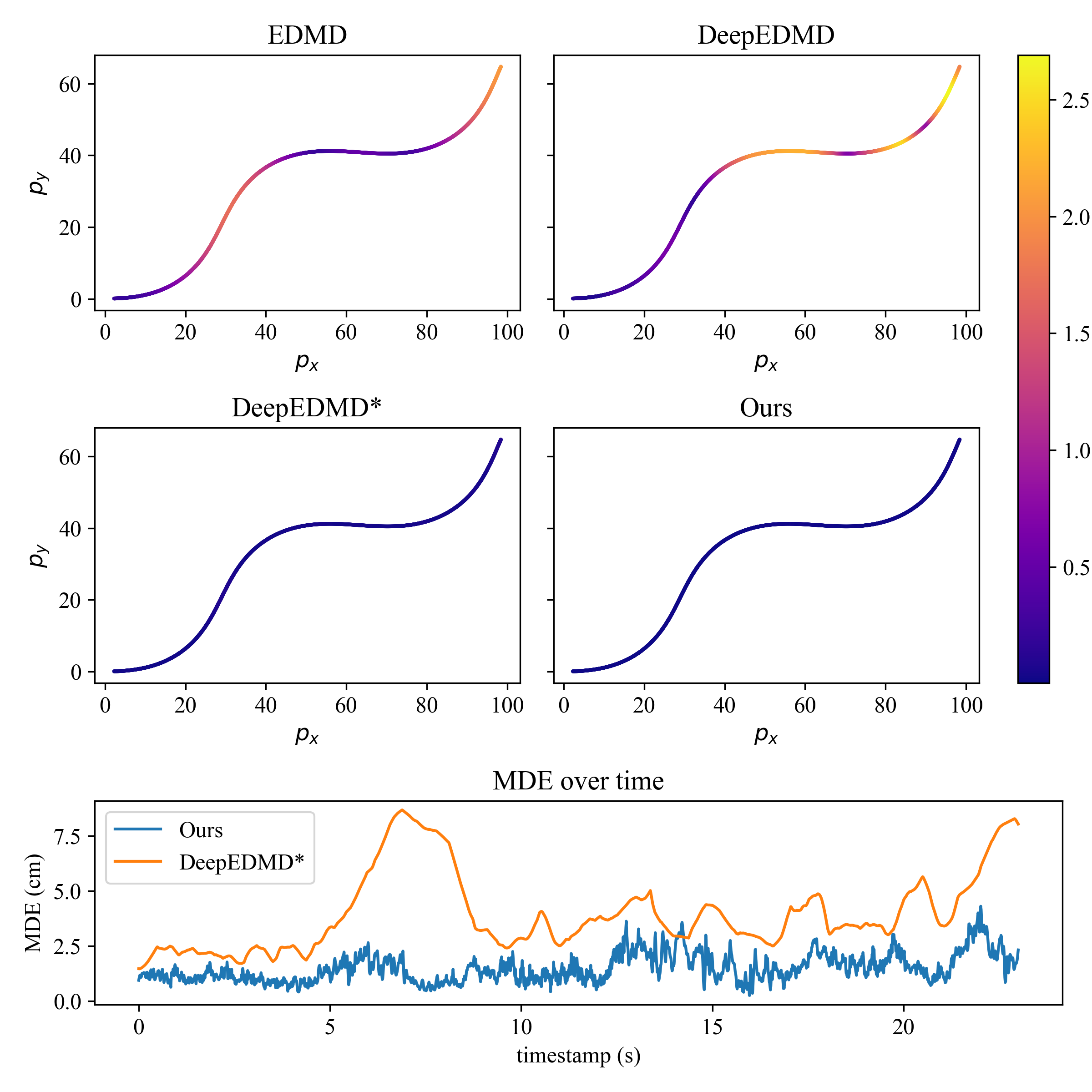}
    \caption{The average trajectory errors for 100-step estimation by our method, DeepEDMD \cite{DeepEDMD}, DeepEDMD* \cite{DeepEDMD}, and EDMD \cite{EDMD}. The value of each point in the trajectory is the average trajectory error of the estimation at this point. We also visualize the average trajectory error of our method and DeepEDMD* \cite{DeepEDMD} in each timestamp for clear visualization.}
    \label{fig:mde}
\end{figure}

Fig. \ref{fig:mde} visualizes the per-timestamp MDE metric for a representative run. EDMD and original DeepEDMD exhibit substantial performance degradation during steering maneuvers, reaching meter-level errors. By contrast, DeepEDMD* and our method maintain centimeter-level accuracy. Our approach consistently outperforms DeepEDMD* at every timestamp, particularly under high-speed steering.

\begin{table}[!t]
	\centering
	\caption{The 100-step dynamics state estmation performance of Koopman-based methods with different encodes (meter).}
    \renewcommand\arraystretch{1.5}
	\begin{threeparttable}
    \resizebox{\linewidth}{!}{
		\begin{tabular}{r | r r | r | r r }
			\toprule
			{Encoder} & D & $H_p$ & {Params.} & MDE $\downarrow$ & FDE $\downarrow$ \\
			\midrule
            Invquad \cite{EDMD} & 16 & 0 & 0.0001 M & 0.5357 & 1.1353 \\
            \hline
            MLP \cite{DeepEDMD} & 16 & 0 & 0.5845 M & 0.0153 & 0.0391 \\
            KAN \cite{liu2025kan} & 16 & 0 & 0.6010 M & 0.0252 & 0.0686  \\
            \hline
            (a) & 16 & 0 & 0.5503 M & 0.0139 & 0.0343  \\
            (b) & 16 & 0 & 0.0022 M & 0.0708 & 0.1919  \\
            \rowcolor{gray!20}{Ours (a+b)} & 16 & 0 & 0.5525 M & 0.0125 & 0.0323  \\
            \rowcolor{gray!20} & 16 & 1 & 0.5544 M & 0.0092 & 0.0240  \\
            \rowcolor{gray!20} & 16 & 5 & 0.5618 M & \textbf{0.0068} & 0.0196  \\
            \rowcolor{gray!20} & 16 & 10 & 0.5711 M & 0.0109 & 0.0263 \\
            \rowcolor{gray!20} & 16 & 15 & 0.5804 M & 0.0097 & 0.0252 \\
            \rowcolor{gray!20} & 64 & 0 & 1.3653 M & 0.0106 & 0.0259   \\
            \rowcolor{gray!20} & 128 & 0 & 1.9713 M & 0.0069 & \textbf{0.0175}\\
			\bottomrule
		\end{tabular}
    }
	\end{threeparttable}
	\label{table:ab-encoder}
\end{table}

\subsection{Ablation Analysis}

To better understand the contribution of each contribution in our proposed method, we conduct comprehensive ablation studies through a series of experiments.

\subsubsection{Effectiveness of dual-branch encoder} 
In this work, we propose a dual-branch encoder that lifts dynamics states to high-dimensional dynamics embeddings for Koopman operator approximation. The encoder comprises a Transformer-based branch (a) and a MLP branch (b). 

To validate the effectiveness of our proposed encoder, we compare it against manual kernel functions \cite{EDMD}, MLP networks \cite{DeepEDMD}, and KAN networks \cite{liu2025kan}. The MLP and KAN baselines are configured with comparable parameter counts and network depths. As shown in Tab. \ref{table:ab-encoder}, all three learnable neural encoders achieve substantially lower long-term estimation errors compared to manual kernel functions \cite{EDMD}, demonstrating the advantage of data-driven encoding over human-designed alternatives. Moreover, our proposed dual-branch encoder outperforms both manual kernels \cite{EDMD} and existing network architectures \cite{DeepEDMD,liu2025kan}, validating its effectiveness.

We further investigate the contribution of each branch in our proposed encoder. As shown in Tab. \ref{table:ab-encoder}, the Transformer-based branch (a) alone achieves satisfactory performance, which can be further enhanced by incorporating a shallow MLP branch (b) that functions as a skip connection—conceptually similar to deep residual learning \cite{resnet}. These results validate the rationality of our dual-branch encoder design.

Theoretically, the Koopman operator requires infinite-dimensional embeddings to exactly capture nonlinear dynamics. In practice, however, finite-dimensional approximations are necessary. To assess the influence of embedding dimension, we compare variants with higher-dimensional embeddings against our default setting. Results in Tab. \ref{table:ab-encoder} indicate that higher dimensions yield better accuracy. For example, the 128-dimensional variant reduces errors by about 45\% relative to $D=16$. For efficiency and fair comparison with existing methods, we retain $D=16$ as the default dimension.

Our proposed encoder is theoretically capable of processing sequential historical dynamics states as input. We therefore investigate the impact of historical state length on modeling performance. The results in Tab. \ref{table:ab-encoder} demonstrate that incorporating historical states enriches the dynamics embeddings and improves estimation accuracy. However, an appropriate length is necessary. As $H$ increases, performance initially improves but begins to degrade once $H$ exceeds 5, likely due to the accumulation of noise from distant historical states.

\begin{table*}[!t]
	\centering
	\caption{The ablation experiments on dynamics state estimation about physics-informed loss and mixture-of-Koopman.}
    \renewcommand\arraystretch{1.5}
	\begin{threeparttable}
    \resizebox{\linewidth}{!}{
		\begin{tabular}{r | r r | r r r r | r r | r r }
			\toprule
			{Model} & PiKo & MoK & MDE@1s $\downarrow$ & FDE@1s $\downarrow$ & MAE@1s $\downarrow$ & FAE@1s $\downarrow$ & MDE@2s $\downarrow$ & MAE@2s $\downarrow$ & MDE@3s $\downarrow$ & MAE@3s $\downarrow$ \\
			\midrule
            MLP \cite{DeepEDMD} & \ding{56} & \ding{56} & 0.0153 m & 0.0391 m & 0.2883$\degree$ & 0.6519$\degree$ & 0.0542 m & 0.7256$\degree$ & 0.1430 m & 1.3227$\degree$ \\
            MLP \cite{DeepEDMD} & \checkmark & \ding{56} & 0.0126 m &  0.0348 m &  0.2651$\degree$ &  0.6481$\degree$ & 0.0528 m & 0.7331$\degree$ & 0.1551 m & 1.2593$\degree$\\
             & & & \textcolor{green}{17.6\%$\downarrow$} & \textcolor{green}{11.0\%$\downarrow$} & \textcolor{green}{8.0\%$\downarrow$} & \textcolor{green}{0.6\%$\downarrow$} & \textcolor{green}{2.6\%$\downarrow$} & \textcolor{red}{1.0\%$\uparrow$} & \textcolor{red}{8.5\%$\uparrow$} & \textcolor{green}{4.8\%$\downarrow$} \\
            \hline
            KAN \cite{liu2025kan} & \ding{56} & \ding{56} & 0.0252 m & 0.0686 m & 0.2626$\degree$ & 0.6017$\degree$ & 0.1100 m & 0.7290$\degree$ & 0.3464 m & 1.6376$\degree$ \\
            KAN \cite{liu2025kan} & \checkmark & \ding{56} & 0.0217 m &  0.0587 m & 0.1614$\degree$ &  0.3828$\degree$ & 0.0894 m & 0.5209$\degree$ & 0.2411 m & 1.2328$\degree$ \\
             & & & \textcolor{green}{17.6\%$\downarrow$} & \textcolor{green}{13.9\%$\downarrow$} & \textcolor{green}{14.4\%$\downarrow$} & \textcolor{green}{36.4\%$\downarrow$} & \textcolor{green}{18.7\%$\downarrow$} & \textcolor{green}{28.5\%$\downarrow$} & \textcolor{green}{30.4\%$\downarrow$} & \textcolor{green}{24.7\%$\downarrow$} \\
            \hline
            Ours (a) & \ding{56} & \ding{56} & 0.0125 m & 0.0323 m & 0.2311$\degree$ & 0.5596$\degree$ & 0.0522 m & 0.6525$\degree$ & 0.1876 m & 1.2629$\degree$ \\
            Ours (b) & \checkmark & \ding{56} & 0.0105 m & 0.0269 m & 0.1043$\degree$ & 0.2744$\degree$ & 0.0497 m & 0.3997$\degree$ & 0.2035 m & 1.0959$\degree$ \\
             & & & \textcolor{green}{16.0\%$\downarrow$} & \textcolor{green}{16.7\%$\downarrow$} & \textcolor{green}{54.9\%$\downarrow$} & \textcolor{green}{51.0\%$\downarrow$} & \textcolor{green}{4.8\%$\downarrow$} & \textcolor{green}{38.7\%$\downarrow$} & \textcolor{red}{8.5\%$\uparrow$} & \textcolor{green}{13.2\%$\downarrow$} \\
            Ours (c) & \ding{56} & \checkmark & 0.0121 m & 0.0293 m & 0.1346$\degree$ & 0.2895$\degree$ & 0.0427 m & 0.3025$\degree$ & \textbf{0.1349} m & 0.5418$\degree$ \\
             & & & \textcolor{green}{3.2\%$\downarrow$} & \textcolor{green}{9.3\%$\downarrow$} & \textcolor{green}{41.8\%$\downarrow$} & \textcolor{green}{48.3\%$\downarrow$} & \textcolor{green}{18.2\%$\downarrow$} & \textcolor{green}{53.6\%$\downarrow$} & \textcolor{green}{\textbf{28.1\%$\downarrow$}} & \textcolor{green}{57.1\%$\downarrow$} \\
            Ours (d) & \checkmark & \checkmark & \textbf{0.0088} m & \textbf{0.0228} m & \textbf{0.1013}$\degree$ & \textbf{0.2306}$\degree$ & \textbf{0.0375} m & \textbf{0.2625}$\degree$ & 0.1366 m & \textbf{0.4952}$\degree$ \\
             & & & \textcolor{green}{\textbf{29.6\%$\downarrow$}} & \textcolor{green}{\textbf{29.4\%$\downarrow$}} & \textcolor{green}{\textbf{56.2\%$\downarrow$}} & \textcolor{green}{\textbf{58.8\%$\downarrow$}} & \textcolor{green}{\textbf{28.2\%$\downarrow$}} & \textcolor{green}{\textbf{59.8\%$\downarrow$}} & \textcolor{green}{27.2\%$\downarrow$} & \textcolor{green}{\textbf{60.8\%$\downarrow$}} \\
			\bottomrule
		\end{tabular}
    }
	\end{threeparttable}
	\label{table:ab-loss-moe}
\end{table*}

\begin{table*}[!t]
	\centering
	\caption{The ablation experiments on dynamics state estimation about MoK operator.}
    \renewcommand\arraystretch{1.5}
	\begin{threeparttable}
    \resizebox{\linewidth}{!}{
		\begin{tabular}{r | r r | r r | r r | r r | r r | r r }
			\toprule
			\multirow{2}{*}{Model} & \multicolumn{2}{c | }{\textbf{c1: FRO}} & \multicolumn{2}{c|}{\textbf{c2: COR}} & \multicolumn{2}{c|}{\textbf{c3: RWL}}  & \multicolumn{2}{c|}{\textbf{c4: LAM}} & \multicolumn{2}{c|}{\textbf{c5: CRM}} & \multicolumn{2}{c}{\textbf{All}} \\
             & MDE $\downarrow$ & FDE $\downarrow$ & MDE $\downarrow$ & FDE $\downarrow$ & MDE $\downarrow$ & FDE $\downarrow$ & MDE $\downarrow$ & FDE $\downarrow$ & MDE $\downarrow$ & FDE $\downarrow$ & MDE $\downarrow$ & FDE $\downarrow$\\
            \midrule
            vanilla & 0.0126 & 0.0324 & \textbf{0.0156} & \textbf{0.0318} & 0.0122 & 0.0318 & 0.0080 & 0.0166 & 0.0070 & 0.0187 & 0.0105 & 0.0269 \\
            MoK & \textbf{0.0120} & \textbf{0.0323} & 0.0186 & 0.0331 & \textbf{0.0089} & \textbf{0.0254} & \textbf{0.0051} & \textbf{0.0113} & \textbf{0.0055} & \textbf{0.0133} & \textbf{0.0088} & \textbf{0.0228} \\
			\bottomrule
		\end{tabular}
    }
	\end{threeparttable}
	\label{table:ab-moe-weight}
\end{table*}

\subsubsection{Effectiveness of physics-informed supervision} 

In this work, we introduce a physics-informed loss function that enforces temporal geometric consistency in vehicle motion, enabling the dynamics model to better learn the underlying physical relationships among different states. To validate its effectiveness, we conduct ablation experiments comparing several Koopman-based methods with and without this supervisory signal. As reported in Tab. \ref{table:ab-loss-moe}, the proposed physics-informed supervision consistently improves modeling performance across different neural network-based encoders, including MLP \cite{DeepEDMD}, KAN \cite{liu2025kan}, and our dual-branch encoder. This suggests that the loss captures general dynamical principles that are independent of network architecture. Furthermore, results in Tabs. \ref{table:main-traj} and \ref{table:main-yaw} demonstrate that incorporating this loss enhances accuracy across most driving patterns. Notably, this improvement is achieved without requiring additional data, network parameters, or training overhead, positioning it as a general enhancement applicable to any Koopman-based dynamics modeling framework.

\subsubsection{Effectiveness of MoK operator}

Given the high-dimensional control space of DETs, practical applications typically employ distinct driving patterns (Fig. \ref{fig:pattern}) to avoid generating infeasible control signals. In this work, we propose a MoK operator as an alternative to the conventional single Koopman operator to better accommodate this multi-pattern setting. To demonstrate its effectiveness, we compare variants using the vanilla Koopman operator against those using the proposed MoK operator. As shown in Tab. \ref{table:ab-loss-moe}, comparing (a) vs. (c) and (b) vs. (c) reveals that MoK consistently improves long-term estimation accuracy. Notably, the improvement becomes more pronounced as the estimation horizon extends, from a 3.2\% gain at 100 steps to a 28.1\% gain at 300 steps. Furthermore, results in Tab. \ref{table:ab-moe-weight} confirm that the MoK operator addresses different driving patterns in a divide-and-conquer fashion, thereby enhancing the overall modeling capability.

\begin{table}[!t]
	\centering
	\caption{The dynamics state estimation performance of different methods in real-world vehicle case testing.}
    \renewcommand\arraystretch{1.5}
	\begin{threeparttable}
    \resizebox{\linewidth}{!}{
		\begin{tabular}{r | r r r r }
			\toprule
			{Method} & MDE $\downarrow$ & FDE $\downarrow$ & MAE $\downarrow$ & FAE $\downarrow$\\
			\midrule
            Linear model (LS) & 2.0039 & 3.9822 & 23.2635 & 46.0574 \\
            Linear model (BP) & 2.1753 & 4.3873 & 0.4035 & 1.1294 \\
            ${\text{EDMD}}^*_{\text{Invquad}}$ \cite{EDMD} & 1.2098 & 2.0335 & 2.0702 & 4.4623 \\
            DeepEDMD* \cite{DeepEDMD} & 0.2774 & 0.4937 & 5.7466 & 10.3690 \\
            \rowcolor{gray!20}KODE (Ours) & 0.2508 & 0.4991 & 0.1025 & 0.2125 \\
            \rowcolor{gray!20}KODE-S2R (Ours) & \textbf{0.0722} & \textbf{0.1206} & \textbf{0.1001} & \textbf{0.2011} \\
			\bottomrule
		\end{tabular}
    }
	\end{threeparttable}
	\label{table:real}
\end{table}

\subsection{Real-world Vehicle Case Testing}

We validate our method on real-world data collected from a six-wheel DET in an industrial park. A single trajectory was divided into training (17.6k frames) and test (5.8k frames) sets to evaluate rapid model identification. Comparisons against a physics-based linear model, EDMD \cite{EDMD}, and DeepEDMD* \cite{DeepEDMD} are presented in Tab. \ref{table:real}. The linear model exhibits inherent limitations, while EDMD improves upon it via kernel-based lifting. DeepEDMD*, with its learnable MLP encoder, achieves substantially better accuracy. Our full model, incorporating the dual-branch encoder and physics-informed loss, yields the most precise heading estimates among all methods trained from scratch. However, all methods perform worse than when trained on large-scale simulation data, which we attribute to overfitting given the limited real-world samples.

\begin{figure}[!t]
    \centering
    \includegraphics[width=0.97\linewidth]{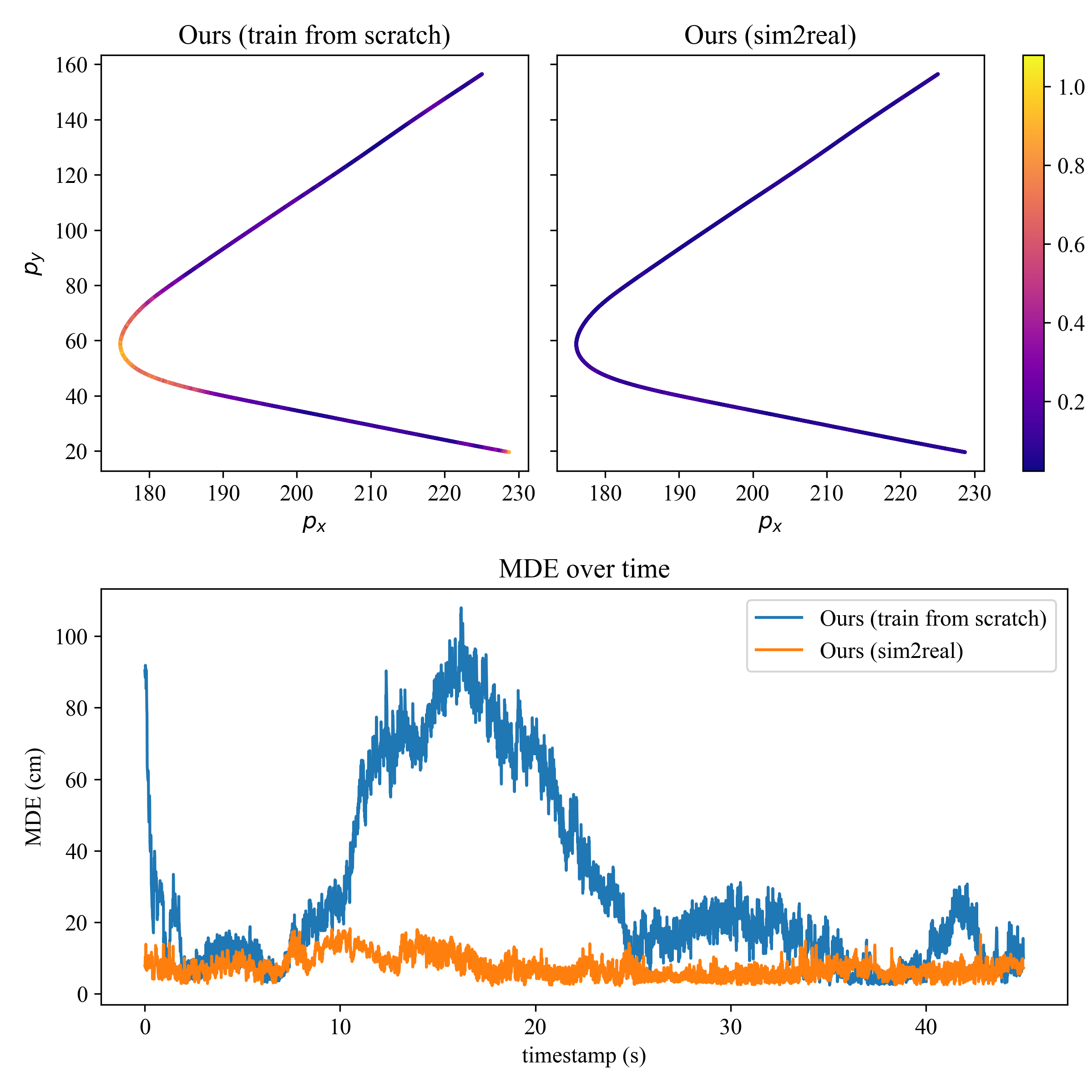}
    \caption{The average trajectory errors for 100-step estimation by our method, both with and without sim-to-real adaptation strategy, across a 45-second clip. The value of each point in the trajectory is the average trajectory error of the estimation at this point. We also visualize the average trajectory error in each timestamp for clear visualization.}
    \label{fig:mde_real}
\end{figure}

Leveraging the insight that the Koopman operator represents the dynamics model parameters (Sec. \ref{sec:transfer}), we test the proposed sim-to-real adaptation strategy (``KODE-S2R'') where the pre-trained encoder is frozen and only the operator is fine-tuned. As shown in Tab. \ref{table:real} and Fig. \ref{fig:mde_real}, this approach effectively overcomes overfitting, demonstrating the efficacy of both the adaptation method and the embeddings learned by our encoder. The adapted model achieves centimeter-level accuracy for 100-step state estimation in real-world conditions.

\section{Conclusions and Future Works}
\label{sec:conclusion}

In this paper, we address the dynamics modeling problem for complex DETs by proposing a novel Koopman-based dynamics model entitled KODE. Koopman operator theory enables the construction of high-precision linear dynamics models within a lifted high-dimensional embedding space. To enhance long-term dynamics state estimation performance, we introduce three main contributions. First, a novel dual-branch encoder is proposed to transform raw dynamics states into expressive embeddings, providing a robust foundation for evolution via the Koopman operator. Second, a physics-informed loss function is derived to enforce geometric consistency in vehicle motion and couple the estimated dynamics states, which can serve as a universal enhancement applicable to any Koopman-based method. Third, to accommodate diverse driving patterns characterized by distinct control input modes, we design a MoK operator that assigns a specialized operator to each pattern, further improving modeling capability. Additionally, efficient adaptation to a new vehicle can be achieved by fine-tuning only the Koopman operator while keeping the encoder frozen. The proposed method is comprehensively evaluated on simulation datasets, demonstrating significant improvements over existing Koopman-based approaches. Real-world vehicle testing further confirms its practical effectiveness, highlighting its strong potential for accurate dynamics modeling in support of high-level autonomous driving.

Future work will focus on integrating the proposed dynamics model into practical closed-loop control systems to validate its effectiveness in real-world autonomous driving scenarios.


\begin{thebibliography}{10}

\bibitem{Jin2019}
X.~Jin, G.~Yin, and N.~Chen, ``Advanced estimation techniques for vehicle system dynamic state: A survey,'' {\em Sensors}, vol.~19, no.~19, 2019.

\bibitem{zhang2024}
T.~Zhang, Y.~Sun, Y.~Wang, B.~Li, Y.~Tian, and F.-Y. Wang, ``A survey of vehicle dynamics modeling methods for autonomous racing: Theoretical models, physical/virtual platforms, and perspectives,'' {\em IEEE Transactions on Intelligent Vehicles}, vol.~9, no.~3, pp.~4312--4334, 2024.

\bibitem{Moser2018}
D.~Moser, R.~Schmied, H.~Waschl, and L.~del Re, ``Flexible spacing adaptive cruise control using stochastic model predictive control,'' {\em IEEE Transactions on Control Systems Technology}, vol.~26, no.~1, pp.~114--127, 2018.

\bibitem{Chen2019}
J.~Chen, W.~Zhan, and M.~Tomizuka, ``Autonomous driving motion planning with constrained iterative lqr,'' {\em IEEE Transactions on Intelligent Vehicles}, vol.~4, no.~2, pp.~244--254, 2019.

\bibitem{spielberg2019neural}
N.~A. Spielberg, M.~Brown, N.~R. Kapania, J.~C. Kegelman, and J.~C. Gerdes, ``Neural network vehicle models for high-performance automated driving,'' {\em Science robotics}, vol.~4, no.~28, p.~eaaw1975, 2019.

\bibitem{guo2024}
J.~Guo, Z.~Dai, M.~Liu, Z.~Xie, Y.~Jiang, H.~Yang, and D.~Xie, ``Distributed drive electric vehicle handling stability coordination control framework based on adaptive model predictive control,'' {\em Sensors}, vol.~24, no.~15, 2024.

\bibitem{Kapania}
N.~R. Kapania, J.~Subosits, and J.~Christian~Gerdes, ``A sequential two-step algorithm for fast generation of vehicle racing trajectories,'' {\em Journal of Dynamic Systems, Measurement, and Control}, vol.~138, p.~091005, 06 2016.

\bibitem{li2015}
L.~Li, G.~Jia, J.~Chen, H.~Zhu, D.~Cao, and J.~Song, ``A novel vehicle dynamics stability control algorithm based on the hierarchical strategy with constrain of nonlinear tyre forces,'' {\em Vehicle System Dynamics}, vol.~53, pp.~1--24, 04 2015.

\bibitem{Tekin2010}
G.~Tekin and Y.~S. Unlusoy, ``Design and simulation of an integrated active yaw control system for road vehicles,'' {\em International Journal of Vehicle Design - INT J VEH DES}, vol.~52, 11 2010.

\bibitem{Azizul2024}
M.~A. Azizul, F.~Ahmad, J.~Karjanto, M.~Che~Hasan, and S.~Sulaiman, ``Modelling, simulation and testing of steer-by-wire system with variable steering ratio control strategy in 14-dof full vehicle model,'' {\em International Journal of Automotive and Mechanical Engineering}, vol.~21, pp.~11784--11808, 11 2024.

\bibitem{Pacejka01011992}
H.~B. Pacejka and E.~Bakker, ``The magic formula tyre model,'' {\em Vehicle System Dynamics}, vol.~21, no.~sup001, pp.~1--18, 1992.

\bibitem{Subosits2021}
J.~K. Subosits and J.~C. Gerdes, ``Impacts of model fidelity on trajectory optimization for autonomous vehicles in extreme maneuvers,'' {\em IEEE Transactions on Intelligent Vehicles}, vol.~6, no.~3, pp.~546--558, 2021.

\bibitem{Vicente2021}
B.~A.~H. Vicente, S.~S. James, and S.~R. Anderson, ``Linear system identification versus physical modeling of lateral–longitudinal vehicle dynamics,'' {\em IEEE Transactions on Control Systems Technology}, vol.~29, no.~3, pp.~1380--1387, 2021.

\bibitem{Johan2019}
J.~Schoukens and L.~Ljung, ``Nonlinear system identification: A user-oriented road map,'' {\em IEEE Control Systems Magazine}, vol.~39, no.~6, pp.~28--99, 2019.

\bibitem{Chrosniak2024}
J.~Chrosniak, J.~Ning, and M.~Behl, ``Deep dynamics: Vehicle dynamics modeling with a physics-constrained neural network for autonomous racing,'' {\em IEEE Robotics and Automation Letters}, vol.~9, no.~6, pp.~5292--5297, 2024.

\bibitem{Xu2019}
J.~Xu, Q.~Luo, K.~Xu, X.~Xiao, S.~Yu, J.~Hu, J.~Miao, and J.~Wang, ``An automated learning-based procedure for large-scale vehicle dynamics modeling on baidu apollo platform,'' in {\em Proceedings of 2019 IEEE/RSJ International Conference on Intelligent Robots and Systems (IROS)}, pp.~5049--5056, 2019.

\bibitem{Mauroy2020}
A.~Mauroy, I.~Mezic, and Y.~Susuki, {\em The Koopman Operator in Systems and Control Concepts, Methodologies, and Applications: Concepts, Methodologies, and Applications}.
\newblock 01 2020.

\bibitem{SCHMID2010}
P.~J. SCHMID, ``Dynamic mode decomposition of numerical and experimental data,'' {\em Journal of Fluid Mechanics}, vol.~656, p.~5–28, 2010.

\bibitem{EDMD}
M.~Korda and I.~Mezić, ``Linear predictors for nonlinear dynamical systems: Koopman operator meets model predictive control,'' {\em Automatica}, vol.~93, pp.~149--160, 2018.

\bibitem{DeepEDMD}
Y.~Xiao, X.~Zhang, X.~Xu, X.~Liu, and J.~Liu, ``Deep neural networks with koopman operators for modeling and control of autonomous vehicles,'' {\em IEEE Transactions on Intelligent Vehicles}, vol.~8, no.~1, pp.~135--146, 2023.

\bibitem{DDK}
Y.~Xiao, X.~Zhang, X.~Xu, Y.~Lu, and J.~Lil, ``Ddk: A deep koopman approach for longitudinal and lateral control of autonomous ground vehicles,'' in {\em 2023 IEEE International Conference on Robotics and Automation (ICRA)}, pp.~975--981, 2023.

\bibitem{Subosits2019}
J.~K. Subosits and J.~C. Gerdes, ``From the racetrack to the road: Real-time trajectory replanning for autonomous driving,'' {\em IEEE Transactions on Intelligent Vehicles}, vol.~4, no.~2, pp.~309--320, 2019.

\bibitem{Timings2013}
J.~P. Timings and D.~J. Cole, ``Minimum maneuver time calculation using convex optimization,'' {\em Journal of Dynamic Systems, Measurement, and Control}, vol.~135, p.~031015, 03 2013.

\bibitem{Liu2018}
K.~Liu, J.~Gong, A.~Kurt, H.~Chen, and U.~Ozguner, ``Dynamic modeling and control of high-speed automated vehicles for lane change maneuver,'' {\em IEEE Transactions on Intelligent Vehicles}, vol.~3, no.~3, pp.~329--339, 2018.

\bibitem{Kabzan2019}
J.~Kabzan, L.~Hewing, A.~Liniger, and M.~N. Zeilinger, ``Learning-based model predictive control for autonomous racing,'' {\em IEEE Robotics and Automation Letters}, vol.~4, no.~4, pp.~3363--3370, 2019.

\bibitem{COSTA2023104469}
G.~Costa, J.~Pinho, M.~A. Botto, and P.~U. Lima, ``Online learning of mpc for autonomous racing,'' {\em Robotics and Autonomous Systems}, vol.~167, p.~104469, 2023.

\bibitem{Hermansdorfer2020}
L.~Hermansdorfer, R.~Trauth, J.~Betz, and M.~Lienkamp, ``End-to-end neural network for vehicle dynamics modeling,'' in {\em Proceedings of 2020 6th IEEE Congress on Information Science and Technology (CiSt)}, pp.~407--412, 2020.

\bibitem{Cao2021}
X.~Cao, H.~Li, C.~Liu, and C.~Qiu, ``Vehicle longitudinal and lateral dynamics modeling by deep neural network,'' in {\em Proceedings of 2021 IEEE International Conference on Real-time Computing and Robotics (RCAR)}, pp.~7--13, 2021.

\bibitem{Kim2022}
T.~Kim, H.~Lee, and W.~Lee, ``Physics embedded neural network vehicle model and applications in risk-aware autonomous driving using latent features,'' in {\em Proceedings of 2022 IEEE/RSJ International Conference on Intelligent Robots and Systems (IROS)}, pp.~4182--4189, 2022.

\bibitem{DRC-baidu}
S.~Jiang, W.~Lin, Y.~Cao, Y.~Wang, J.~Miao, and Q.~Luo, ``Learning-based vehicle dynamics residual correction model for autonomous driving simulation,'' in {\em Proceedings of 2021 IEEE International Intelligent Transportation Systems Conference (ITSC)}, pp.~782--789, 2021.

\bibitem{DRC-baidu2}
S.~Jiang, Y.~Wang, W.~Lin, Y.~Cao, L.~Lin, J.~Miao, and Q.~Luo, ``A high-accuracy framework for vehicle dynamic modeling in autonomous driving,'' in {\em Proceedings of 2021 IEEE/RSJ International Conference on Intelligent Robots and Systems (IROS)}, pp.~6680--6687, 2021.

\bibitem{miao2025}
J.~Miao, R.~Yan, B.~Zhang, T.~Wen, J.~Li, Z.~Fu, K.~Jiang, M.~Yang, J.~Huang, Z.~Zhong, and D.~Yang, ``Residual learning towards high-fidelity vehicle dynamics modeling with transformer,'' {\em IEEE Robotics and Automation Letters}, vol.~10, no.~7, pp.~7404--7411, 2025.

\bibitem{Williams2015}
M.~O. Williams, I.~G. Kevrekidis, and C.~W. Rowley, ``A data–driven approximation of the {K}oopman operator: {E}xtending dynamic mode decomposition,'' {\em Journal of Nonlinear Science}, vol.~25, pp.~1307--1346, Dec. 2015.

\bibitem{Mamakoukas2021}
G.~Mamakoukas, M.~L. Castaño, X.~Tan, and T.~D. Murphey, ``Derivative-based koopman operators for real-time control of robotic systems,'' {\em IEEE Transactions on Robotics}, vol.~37, no.~6, pp.~2173--2192, 2021.

\bibitem{Matthew2015}
M.~O. Williams, C.~W. Rowley, and I.~G. Kevrekidis, ``A kernel-based method for data-driven koopman spectral analysis,'' 2015.

\bibitem{Lusch2018}
B.~Lusch, J.~N. Kutz, and S.~L. Brunton, ``Deep learning for universal linear embeddings of nonlinear dynamics,'' {\em Nature Communications}, vol.~9, no.~1, p.~4950, 2018.

\bibitem{SafEDMD}
R.~Strässer, M.~Schaller, K.~Worthmann, J.~Berberich, and F.~Allgöwer, ``Safedmd: A koopman-based data-driven controller design framework for nonlinear dynamical systems,'' {\em Automatica}, vol.~185, p.~112732, 2026.

\bibitem{DK-U}
H.~Shi and M.~Q.-H. Meng, ``Deep koopman operator with control for nonlinear systems,'' {\em IEEE Robotics and Automation Letters}, vol.~7, no.~3, pp.~7700--7707, 2022.

\bibitem{Takeishi2017}
N.~Takeishi, Y.~Kawahara, and T.~Yairi, ``Learning koopman invariant subspaces for dynamic mode decomposition,'' in {\em Proceedings of the 31st International Conference on Neural Information Processing Systems}, NIPS'17, (Red Hook, NY, USA), p.~1130–1140, Curran Associates Inc., 2017.

\bibitem{Brunton2022}
S.~L. Brunton, M.~Budi\v{s}i\'{c}, E.~Kaiser, and J.~N. Kutz, ``Modern koopman theory for dynamical systems,'' {\em SIAM Review}, vol.~64, no.~2, pp.~229--340, 2022.

\bibitem{ESO-DK}
H.~Chen and C.~Lv, ``Incorporating eso into deep koopman operator modeling for control of autonomous vehicles,'' {\em IEEE Transactions on Control Systems Technology}, vol.~32, no.~5, pp.~1854--1864, 2024.

\bibitem{tancik2020fourier}
M.~Tancik, P.~Srinivasan, B.~Mildenhall, S.~Fridovich-Keil, N.~Raghavan, U.~Singhal, R.~Ramamoorthi, J.~Barron, and R.~Ng, ``Fourier features let networks learn high frequency functions in low dimensional domains,'' {\em Advances in neural information processing systems}, vol.~33, pp.~7537--7547, 2020.

\bibitem{gelu}
D.~Hendrycks and K.~Gimpel, ``Gaussian error linear units (gelus),'' {\em arXiv: Learning}, 2016.

\bibitem{vit}
A.~Dosovitskiy, L.~Beyer, A.~Kolesnikov, D.~Weissenborn, X.~Zhai, T.~Unterthiner, M.~Dehghani, M.~Minderer, G.~Heigold, S.~Gelly, J.~Uszkoreit, and N.~Houlsby, ``An image is worth 16x16 words: Transformers for image recognition at scale,'' in {\em 9th International Conference on Learning Representations, {ICLR} 2021, Virtual Event, Austria, May 3-7, 2021}, OpenReview.net, 2021.

\bibitem{poet}
J.~Miao, K.~Jiang, Y.~Wang, T.~Wen, Z.~Xiao, Z.~Fu, M.~Yang, M.~Liu, J.~Huang, Z.~Zhong, and D.~Yang, ``Poses as queries: End-to-end image-to-lidar map localization with transformers,'' {\em IEEE Robotics and Automation Letters}, vol.~9, no.~1, pp.~803--810, 2024.

\bibitem{moe}
W.~Gan, Z.~Ning, Z.~Qi, and P.~S. Yu, ``Mixture of experts (moe): A big data perspective,'' {\em Information Fusion}, vol.~127, p.~103664, 2026.

\bibitem{vaswani2017attention}
A.~Vaswani, ``Attention is all you need,'' {\em Proceedings of Advances in Neural Information Processing Systems (NeurIPS)}, 2017.

\bibitem{liu2025kan}
Z.~Liu, Y.~Wang, S.~Vaidya, F.~Ruehle, J.~Halverson, M.~Soljacic, T.~Y. Hou, and M.~Tegmark, ``{KAN}: Kolmogorov{\textendash}arnold networks,'' in {\em The Thirteenth International Conference on Learning Representations}, 2025.

\bibitem{resnet}
K.~He, X.~Zhang, S.~Ren, and J.~Sun, ``Deep residual learning for image recognition,'' in {\em Proceedings of 2016 IEEE Conference on Computer Vision and Pattern Recognition (CVPR)}, pp.~770--778, 2016.

\end{thebibliography}





\vfill

\end{document}